\pdfoutput=1
\documentclass[11pt]{article}

\usepackage{acl}
\usepackage{authblk}

\usepackage{times}
\usepackage{latexsym}

\usepackage[T1]{fontenc}
\usepackage[utf8]{inputenc}

\usepackage{microtype}

\usepackage{array}
\usepackage{tabularx}
\usepackage{booktabs}
\usepackage{graphicx}
\usepackage{threeparttable}
\usepackage{multirow}
\usepackage{amsmath}
\usepackage{subcaption}
\usepackage{xurl}
\usepackage{xspace}

\providecommand{\baseline}{FAFT\xspace}

\title{Transformers with Learnable Activation Functions}

\author[1]{Haishuo Fang}
\author[1]{Ji-Ung Lee}
\author[1,2]{Nafise Sadat Moosavi}
\author[1]{Iryna Gurevych}
 
\affil[1]{Ubiquitous Knowledge Processing Lab (UKP Lab)\protect\\ Department of Computer Science and Hessian Center for AI (hessian.AI)\protect\\ Technical University of Darmstadt \protect\\ \url{www.ukp.tu-darmstadt.de}\protect\vspace{.6em}}
\affil[2]{Department of Computer Science, The University of Sheffield}

 \date{}

\begin{document}
\maketitle

\begin{abstract}
Activation functions can have a significant impact on reducing the topological complexity of input data and therefore, improving a model's performance. However, the choice of activation functions is seldom discussed or explored in Transformer-based language models.
As a common practice, commonly used activation functions like Gaussian Error Linear Unit (GELU) are chosen beforehand and then remain fixed from pre-training to fine-tuning.
In this paper, we investigate the impact of activation functions on Transformer-based models by utilizing rational activation functions (RAFs). 
In contrast to fixed activation functions (FAF), RAFs are capable of learning the optimal activation functions from data. 
Our experiments show that the RAF-based Transformer model (RAFT) achieves a better performance than its FAF-based counterpart (\baseline). 
For instance, we find that RAFT outperforms \baseline on the GLUE benchmark by 5.71 points when using only 100 training examples and by 2.05 points on SQuAD with all available data.
Analyzing the shapes of the learned RAFs further unveils that they vary across different layers and different tasks; opening a promising way to better analyze and understand large, pre-trained language models.\footnote{Code, models, and datasplits are available on GitHub \url{https://github.com/UKPLab/2022-RAFT}.} 
\end{abstract}

\section{Introduction}
Activation functions introduce non-linearity and increase neural networks' representational capacity, and therefore, play an essential role in designing deep learning models \cite{nwankpa2018activation,sharma2017activation,dubey2022activation}.
\citet{naitzat2020topology} explain the importance of activation functions by proposing to consider data as a topology with its own shape. 
They empirically show that activation functions accelerate the data topology transformation through different layers of a neural network to simplify its complexity and make it linearly separable in the output space.
Their experiments show that choosing the right activation function can have a significant impact on the overall performance.

While any activation function can be used with Transformers~\cite{vaswani2017attention}, their choice is made before pre-training and remains fixed afterwards. 
Hence, the inductive bias an activation function imposes on the model cannot be adjusted during pre-training or fine-tuning.
As many Transformer-based models are pre-trained on a large amount of data, and changing the activation function for or during fine-tuning may negatively impact the performance\footnote{In our preliminary experiments, the performance of BERT becomes worse on downstream tasks when the activation functions are changed after pre-training.}.
Moreover, the simple case of finding the optimal combination of $k$ different activation functions in $n$ different feedforward layers results in $k^n$ possible combinations and becomes intractable; e.g., 531,441 experiments for a 12-layer BERT model and three different activation functions.
As a result, most Transformer-based pre-trained models adopt the GELU activation function that has been initially used for the BERT model~\cite{devlin2018bert}.

To overcome the limitation of using a potentially suboptimal activation function that remains fixed during training, we propose to use a learnable activation function, namely, the rational activation function (RAF, \citealt{Molina2020Pade}).
The RAF is a universal function approximator that can approximate any existing activation function.
The advantage of using RAFs over fixed activation functions (FAF) such as ReLU or GELU, is that the model can learn the optimal activation function from the data during (pre)training without the need to consider the choice of activation function as an additional dimension during hyperparameter tuning.\footnote{\citet{liu2018darts} consider different activation functions during Neural Architecture Search~\citep{zoph2017neural}, but this becomes quickly infeasible for compute-intensive experiments such as pre-training large language models.}
To evaluate the effectiveness of RAFs, we pre-train two encoder-only Transformers using RAF and GELU respectively, within an academic budget. 
In our experiments, we find that:
\begin{itemize}
    \item The RAF-based Transformer (RAFT) learns different activation functions at different layers after pre-training with shapes that differ from frequently used activation functions.
    \item During fine-tuning, RAFT outperforms its fixed activation function counterpart (\baseline) on the general language understanding benchmark (GLUE) and the SQuAD machine reading comprehension dataset in various settings. 
    \item After fine-tuning, the learned RAFs of the top layers are more task-specific and change the most, which are corresponding to layer behaviors of Transformers according to prior work~\cite{mosbach2020interplay,merchant2020happens,zhou2021closer}.
    This provides new opportunities to analyze language models with respect to their learned activation functions at different layers for different tasks.
    \item RAFT boosts the performance when combined with a parameter-efficient fine-tuning approach, i.e., BitFit~\cite{zaken2021bitfit}, which improves the model performance by 3.08 points in full-data scenario.
\end{itemize}

\section{Related Work}

\paragraph{Activation functions.}
There exists various predefined activation functions such as Sigmoid,
Hyperbolic Tangent (Tanh), Rectified Linear Unit (ReLU, \citealt{fukushima1969visual}), and Gaussian Error Linear Unit (GELU, \citealt{DBLP:journals/corr/HendrycksG16}).
There are also approaches that leverage automatic search to obtain optimal combinations of several base activation functions in a predefined search space~\cite{ramachandran2017searching,manessi2018learning,sutfeld2020adaptive,bingham2022discovering,bingham2020evolutionary}. 
For instance, \citet{ramachandran2017searching} discovered the Swish activation function by using this method.
\citet{bingham2020evolutionary} show that further extending the search space using evolutionary algorithms can also lead to an improvement. 
Finally, several search-based works investigate how to train a combination of a set of activation functions to better adapt to specific tasks and architectures~\cite{manessi2018learning,sutfeld2020adaptive,bingham2022discovering}. 
One substantial drawback of these search-based methods is that they are computationally expensive. 
Especially for pre-trained language models where pre-training is costly, it is infeasible to perform a hyperparameter search for selecting the best activation function (even more so their combination).
In contrast, the flexibility of rational activation functions (RAFs) allows them to be trained along with the model parameters in an end-to-end fashion~\cite{Molina2020Pade}.
Therefore, they can learn the optimized activation function from data during training.
RAFs have been successfully used in deep reinforcement learning for improving plasticity~\cite{delfosse2021recurrent}, cell detection models in biology \cite{prangemeier2020attention}, and adapter architectures~\cite{moosavi2022adaptable}. 

\begin{table}[htb]
    \centering
    \resizebox{0.8\columnwidth}{!}{
    \begin{tabular}{@{}lr@{}}
    \toprule
    Model & Act. Funct. \\
    \midrule
    BERT~\cite{devlin2018bert} & GELU \\
    GPT-1~\cite{radford2018improving} & GELU \\
    RoBERTa~\cite{liu2019roberta} & GELU \\
    XLNet~\cite{yang2019xlnet} & GELU \\
    ALBERT~\cite{lan2019albert} & GELU \\ 
    GPT-2$^*$~\cite{radford2019language} & GELU \\
    Megatron-LM~\cite{shoeybi2019megatron} & GELU \\
    ELECTRA$^+$~\cite{clark2020electra} & GELU \\
    T5~\cite{raffel2020exploring} & ReLU \\
    T5v1.1~\cite{raffel2020exploring} & GeGLU \\
    DeBERTa$^+$~\cite{he2020deberta} & GELU \\
    BART~\cite{lewis-etal-2020-bart} & GELU \\ 
    GPT-3$^*$~\cite{brown2020language} & GELU \\
    Jurassic$^*$~\cite{lieber2021jurassic} & GELU \\
    Gopher$^*$~\cite{rae2021scaling} & GELU \\
    Megatron-Turing NLG$^*$~\cite{smith2022using} & GELU \\
    Chinchilla$^*$~\cite{hoffmann2022training} & GELU \\
    CANINE$^+$~\cite{clark2022canine} & GELU \\
    LaMBDA~\cite{thoppilan2022lamda} & GeGLU  \\
    OPT~\cite{zhang2022opt} & ReLU \\ 
    \bottomrule
    \end{tabular}
    }
    \caption{Activation functions in different NLP Transformer models. Models marked by $^*$ do not explicitly state the activation function but refer to GPT-1 as the base architecture ($^+$ refers to BERT respectively). GeGLU is a variant that combines GELU and GLU.}
    \label{tab:activation_functions_models}

\end{table}

\paragraph{Frequently used activation functions in NLP.}
Table~\ref{tab:activation_functions_models} shows a list of 20 different language models that have been introduced after BERT. 
As we see, the vast majority of the works (80\%) use the GELU activation function.
Moreover, many works even do not explicitly state the used activation function (45\%). 
There are only a few works that investigate the impact of activation functions on pre-trained Transformer models.
\citet{so2021searching} leverage automatic search methods to identify more efficient Transformer architectures. 
They find that a combination of squared ReLU used in the feedforward network (FFN) layer and a convolution layer added in self-attention can lead to a substantial boost in performance. 
\citet{shazeer2020glu} replace the FFN in the Transformer with a gated linear unit (GLU, \citealt{10.5555/3305381.3305478}) combined with different activation functions and find a higher performance during pre-training as well as on downstream tasks. 
In our work, we do not change the structure of FFNs and only replace activation functions in them. 

Closest to our work is the work by \citet{moosavi2022adaptable} who investigate the use of RAF in adapters~\cite{pmlr-v97-houlsby19a}; i.e., lightweight layers that are added on top of pre-trained Transformer layers.
They propose adaptable adapters that consist of RAFs and learnable switches to select a subset of adapter layers during training.
They show that using both RAFs and a fewer number of adapter layers results in considerable performance gains, especially in low-data settings.
However, only using RAF instead of ReLU does not result in a considerable gain in their experiments. 
Furthermore, adapter layers are only added and updated during fine-tuning, as a result using RAF in adapter layers has a limited impact compared to already applying them for pre-training. 

In this work, we show that using RAF in Transformer layers brings additional flexibility to the model to learn the optimized activation function for each of its layers during training, and that this additional flexibility benefits both pre-training and fine-tuning steps.

\section{RAFT: RAF-based Transformers } \label{sec:approach}

We adopt the BERT architecture~\cite{devlin2018bert} where all activation functions in feed-forward layers $Activation(W_{1}X)W_{2}$ are replaced with rational activation functions (illustrated in Appendix~\ref{sec:model architecture}).
The equation of rational activation function $F(x)$ is as below:
\begin{equation}
    F(x) = \frac{P(x)}{Q(x)} = \frac{\sum_{j=0}^{m}a_{j}x^{j}}{1+|\sum_{k=0}^{n}b_{k}x^{k}|}
\end{equation}
Where $a$ and $b$ are learnable parameters, and $m$ and $n$ are degrees of $F(x)$, which decide the complexity and fitting ability of rational functions.
Following \citet{Molina2020Pade}, we use the \textit{safe PAU} formulation that further stabilizes training.

\paragraph{Selecting $m$ and $n$.} 
Similar to Taylor series, the higher the degrees $m$ and $n$ are, the more precise is the approximation of rational functions.
However, indefinitely increasing the degrees also means adding more complexity and increasing training time.
The challenge is to find suitable degrees that leads to rational functions with a strong fitting ability while keeping their complexity as low as possible. 
As this is still an open question, we set the search space of $m$ and $n$ to $\{4,5\}$, and evaluate their ability to approximate the GELU function in the range of [-3,3]. 
Our results show that using $m=5$ and $n=4$ perfectly fits the GELU function with a low complexity and thus, are adopted in this work (cf. Figure~\ref{fig:approx_gelu}, Appendix~\ref{sec:fitting_abilities}). 
This matches the findings in previous work~\cite{telgarsky2017rational,Molina2020Pade,delfosse2021recurrent} as well.
So overall, each rational activation function adds nine parameters, resulting in a total of 108 additional parameters in a 12-layer Transformer model (less than 0.000098\% of its original parameters).
The weights of $F(x)$ can further be initialized to approximate any existing activation functions. 
In our experiments, we initialize it with weights that approximate GELU.

\section{Pre-training}\label{sec:pretraining}
To evaluate the viability of RAFT, we pre-train two comparable Transformer models from scratch---one using the common fixed GELU activation function (\baseline), and another one using RAFs (RAFT).
%We then further fine-tune and evaluate both models on two NLP benchmarks.

\paragraph{Model architecture.} 
For our experiments, we use a frequently considered model configuration and train 12 Transformer encoder layers with a hidden size of 768 and 12 attention heads~\citep{devlin2018bert,liu2019roberta,rae2021scaling,zhang2022opt}.
The only difference between RAFT and \baseline is the use of RAFs instead of GELUs as activation functions.

\paragraph{Data.} 
We use English Wikipedia as our pre-training data.\footnote{\url{https://dumps.wikimedia.org}}
The dataset consists of $3.8\times10^9$ tokens from which we select 50k sentences containing $6.4\times10^6$ tokens as the validation data. 
% Each sentence is split into character- or word-level tokens with Byte-Pair Encoding \cite{sennrich2015neural}. 
% Following \citet{izsak2021train}, we set the sequence length to 128 tokens throughout the whole pre-training process for a better computational efficiency.

\paragraph{Pre-training objective.}
Following RoBERTa \cite{liu2019roberta}, we use dynamic masked language modeling (MLM) as our learning task and randomly mask tokens in the input sentences at each step before feeding them into the model.
%The masking strategy is the same as RoBERTa.
We use the same masking probabilities and mask 15\% of the tokens with an 80\% chance of replacing them with the $[\mathrm{MASK}]$ token, a 10\% chance of replacing them with a randomly selected different token, and a 10\% chance of not replacing them at all.

\paragraph{Training parameters.} 
As our primary goal is to validate the effectiveness of RAFs in Transformers rather than releasing a RoBERTa-like model, we focus on training two comparable models within a limited training budget.
Both models are optimized using AdamW~\cite{loshchilov2017decoupled} with $\beta_{1}=0.9$, $\beta_{2}=0.98$ and a weight decay of 0.01. 
The learning rate $lr_{\theta}$ is set to 7E\nobreakdash-4 for both models while the learning rate $lr_{\mathrm{RAF}}$ for the RAF coefficients is set to 5E\nobreakdash-3. 
Both learning rates are warmed up over the first 1\% steps, then $lr_{\theta}$ decays linearly while $lr_{\mathrm{RAF}}$ remains constant.\footnote{We find in our preliminary experiments that a constant rational learning rate with warm up leads to better results.} 
The batch size is set to 4096.
Tuning hyperparameters during pre-training is expensive, to conduct hyperparameters tuning of both models with limited resources, we follow up 24hour BERT \cite{izsak2021train} to pre-train the model for 23k steps equipped with various methods to accelerate training, including mixed-precision, sparse output prediction, fused linear layer, and tied embeddings~\cite{press-wolf-2017-using}. Detailed parameters and results of hyperparameter tuning are provided in Appendix~\ref{sec:hyperparam}.
% After 23k steps, validation loss of both models go flat and there is only a slow decrease in the evaluation loss afterwards (cf. Figure~\ref{fig:valid_loss}, Appendix~\ref{sec:hyperparam}).
It takes $\sim$16 hours for RAFT and $\sim$12 hours for \baseline using four A100 GPUs.
% we achieve a $\sim$50\% reduction of pre-training time compa, i.e., $\sim$16 hours for RAFT and $\sim$12 hours for \baseline using four A100 40GiB GPUs.

%\paragraph{Pre-training setup.} 
%Pre-training a high-performance Transformer-based language model from scratch follow RoBERTa setup is prohibitively expensive. For example, training a language model for 500k steps with a batch size of 8k would need approximately 1 month on 4 A100 GPUs even with mixed-precision training.
%In this work, our goal is not to release a RoBERTa-like model but to validate the effectiveness of RAFs in Transformers.
%To conduct this experiment with an academic budget \cite{izsak2021train}, we train the models for 23k steps combined with different acceleration methods, including mixed-precision training, sparse output prediction and tied embeddings \cite{press-wolf-2017-using}. 
%The whole pre-training process is conducted on four A100 40GiB GPUs for both models and takes $\sim\!16$ hours for RAFT and $\sim\!12$ hours for \baseline.
%Exact pre-training details are provided in Appendix~\ref{sec:hyperparam}.

\begin{table}[!t]
\centering
\resizebox{0.8\columnwidth}{!}{%
\begin{tabular}{@{}ccc@{}}
\toprule
Model        & Validation loss & Validation PPL \\ \midrule
\baseline & 1.645           & 5.18           \\
RAFT         & \textbf{1.611}  & \textbf{5.00}  \\ \bottomrule
\end{tabular}%
}
\caption{Performance of the models on the validation set after pre-training.}
\label{tab:pre-train_loss}
\end{table}

\paragraph{Results.} Table~\ref{tab:pre-train_loss} shows the MLM validation losses and validation perplexity of the best performing hyperparameter configuration for RAFT and \baseline.
We observe that RAFT achieves a bit lower perplexity than \baseline during pre-training.
The learned RAFs vary across different layers after pre-training (cf. Figure~\ref{fig:rfs-after-pretraining}, Appendix~\ref{sec:appendix:e}).
More analysis is conducted in Section~\ref{sec:analysis}.
%The pre-trained RAFs could be found in Figure~\ref{fig:rfs-after-pretraining}.

\begin{table*}
\centering

\resizebox{\textwidth}{!}{%
\begin{threeparttable}
\begin{tabular}{cclcccccll}
\hline
Model & ColA & \multicolumn{1}{c}{SST2} & MRPC & QQP & STSB & MNLI-matched/mismatched & QNLI & \multicolumn{1}{c}{RTE} & Avg. \\ \hline
\multicolumn{1}{l}{\textit{low-data 100 examples}\tnote{1}} & \multicolumn{1}{l}{} &  & \multicolumn{1}{l}{} & \multicolumn{1}{l}{} &  &  & \multicolumn{1}{l}{} &  &  \\
\baseline & 1.88±2.27 & \multicolumn{1}{c}{71.02±5.61} & 74.88±0.23 & 55.19±5.96 & 57.57±8.32 & 32.86±1.50/32.92±1.46 & 53.34±3.24 & \multicolumn{1}{c}{\textbf{53.14±1.67}} & \multicolumn{1}{c}{48.07} \\
RAFT$^\mathrm{full}$ & 4.38±3.2 & \multicolumn{1}{c}{\textbf{73.28±3.95}} & \textbf{75.89±1.39} & \textbf{62.65±2.86} & 70.30±3.44 & 38.31±1.87/39.06±2.35 & \textbf{63.58±3.74} & \multicolumn{1}{c}{53.0±1.91} & \multicolumn{1}{c}{53.38} \\
RAFT$^\mathrm{fixed}$ & \textbf{7.25±4.77} & 72.04±5.04 & 75.76±0.65 & 62.15±4.09 & \textbf{71.39±3.56} & \textbf{39.3±1.60/40.4±1.73} & \multicolumn{1}{l}{63.13±3.05} & 52.6±2.99 & \textbf{53.78} \\ \hline
\textit{low-data 300 examples}\tnote{1} & \multicolumn{1}{l}{} &  & \multicolumn{1}{l}{} & \multicolumn{1}{l}{} &  &  & \multicolumn{1}{l}{} &  &  \\
\baseline & \multicolumn{1}{l}{13.12±5.29} & 77.67±3.07 & \multicolumn{1}{l}{\textbf{79.37±1.56}} & 66.63±1.35 & 76.70±1.89 & 43.74±2.20/45.33,2.29 & 69.17±2.25 & 55.45±2.66 & 58.58 \\
RAFT$^\mathrm{full}$  & \multicolumn{1}{l}{12.36±5.07} & 78.22±2.10 & \multicolumn{1}{l}{77.84±1.09} & \textbf{68.25±1.01} & 79.77±2.34 & \textbf{45.70±1.69/47.27±1.86} & 71.92±1.10 & 54.70±2.26 & 59.56 \\
RAFT$^\mathrm{fixed}$ & \multicolumn{1}{l}{\textbf{17.34±3.23}} & \textbf{78.95±2.33} & \multicolumn{1}{l}{76.97±0.96} & 68.20±0.76 & \textbf{80.32±0.1} & 45.35±1.62/46.53±1.63 & \textbf{72.07±1.56} & \textbf{55.78±2.72} & \textbf{60.17} \\ \hline
\textit{Full data}\tnote{2} & \textit{} & \multicolumn{1}{c}{\textit{}} & \textit{} & \textit{} & \textit{} & \textit{} & \textit{} & \multicolumn{1}{c}{\textit{}} & \textit{} \\
\baseline & 43.18±1.52 & \multicolumn{1}{c}{89.2±0.63} & 86.42±1.37 & 88.08±0.08 & \textbf{87.08±0.21} & 80.92±0.21/81.78±0.22 & \multicolumn{1}{l}{\textbf{89.42±0.38}} & 62.22±1.35 & 78.70 \\
RAFT$^\mathrm{full}$ & \textbf{45.84±1.47} & \multicolumn{1}{c}{89.85±0.45} & \textbf{87.21±0.54} & \textbf{88.27±0.10} & 86.96±0.29 & 80.88±0.22/81.85±0.23 & \multicolumn{1}{l}{89.32±0.20} & \textbf{64.44±2.49} & \textbf{79.40} \\ 
RAFT$^\mathrm{fixed}$ & \multicolumn{1}{l}{45.66±1.55} & \textbf{90.06±0.70} & \multicolumn{1}{l}{86.36±1.03} & 88.21±0.06 & 86.64±0.24 & \textbf{81.10±0.22/82.06±0.21} & \multicolumn{1}{l}{89.36±0.34} & 63.90±2.85 & 79.28 \\ \hline
\end{tabular}%
\begin{tablenotes}
\footnotesize
\item[1] Results are averaged over ten random seeds: 5309, 202206, 20220602, 2259, 49, 2022, 1046, 622, 320, 53
\item[2] Results are averaged over six random seeds: 5309, 202206, 20220602, 2259, 49, 2022
\end{tablenotes}
\end{threeparttable}
}
\caption{The performance of RAFT and \baseline on the GLUE benchmark across different data sizes. RAFT$^\mathrm{full}$ fine-tunes all model parameters including RAFs. RAFT$^\mathrm{fixed}$ instead fixes the RAFs pre-training.}
\label{tab:glue}
\end{table*}

\section{Fine-tuning} \label{sec:fine-tuning}
We conduct experiments on the General Language Understanding Evaluation (GLUE) benchmark \cite{wang2018glue} and SQuAD \cite{rajpurkar-etal-2016-squad} to see how well pre-trained RAFs can adapt to downstream tasks.
Dataset descriptions are provided in Appendix~\ref{sec:data}.
We further investigate the flexibility of the pre-trained RAFs by considering different training data sizes especially in a low-data regime. 
We fine-tune RAFT in two different settings:
% (a) RAFT$^\mathrm{full}$ where we fine-tune the whole model, i.e., all model parameters including the RAFs and (b)  RAFT$^\mathrm{fixed}$ where we fix the pre-trained RAFs and only tune the other model parameters.
\begin{itemize}
    \item RAFT$^\mathrm{full}$: We fine-tune the whole model, i.e., all model parameters including the RAFs.
    \item RAFT$^\mathrm{fixed}$: We fix the pre-trained RAFs and only tune the rest of the parameters.
\end{itemize}

\subsection{Evaluation on the GLUE Benchmark} 
We evaluate pre-trained models on GLUE benchmark in different data settings: (a) the full-data scenario, and (b) two low-data scenarios when only 100 or 300 labelled examples are available.

\paragraph{Experimental Setup.}
We split 75\% of the training dataset as the training set and use the remaining 25\% as the development set in the full-data scenario.
Following previous works, we use the provided development set as the test dataset.
For our low-data scenarios, we randomly sample 100 or 300 examples with ten different random seeds and report the average and standard deviation across all runs. 
For the full-data scenario, we report the average and standard deviation of the results across six runs with different random seeds. 
We use the same evaluation metrics as proposed in the GLUE benchmark; more specifically, for MRPC, QQP, and STSB, we use the average of the two corresponding metrics as the final score.\footnote{Note that the full-data scenario is computationally more expensive to run, but also more stable as the training instances experience less variability.}

% \paragraph{Hyperparameter Tuning.} 
% The hyperparameter search space could be found in Appendix .

\paragraph{Results.} Table~\ref{tab:glue} shows the performance of RAFT and \baseline on the GLUE benchmark. 
We observe that on average, RAFT achieves consistent improvements in all data settings. 
We further find that especially in the low-data scenarios, the flexible activation functions of RAFT substantially outperform their static GLUE counterparts of the \baseline model. 
For 100 examples, RAFT achieves better results in seven out of eight tasks, outperforming \baseline by 5.31 points (RAFT$^\mathrm{full}$) and 5.71 points (RAFT$^\mathrm{fixed}$) on average, respectively.
While the performance gap becomes smaller as the number of examples increases, the tendency remains the same with an average performance gain of 0.98 points (RAFT$^\mathrm{full}$) and 1.59 points (RAFT$^\mathrm{fixed}$) for 300 examples.
In the full data scenario, RAFT still outperforms \baseline by 0.7 (RAFT$^\mathrm{full}$) and 0.58 (RAFT$^\mathrm{fixed}$) points on average.

Our experiments indicate that fixing the RAFs is a better choice for the GLUE benchmark in the low-data scenarios.
We conjecture that one reason for this may be that the number of instances to tune all parameters of the model are insufficient. 
On the contrary, we find that in the full-data scenario tuning RAFs can lead to better results. 
The increasing number of instances especially benefit RAFs as they can better adapt to different downstream tasks and learn better features. 
We provide further analysis in Section~\ref{sec:analysis}.

\subsection{Evaluation on SQuAD} 

Similar to GLUE, we evaluate models on SQuAD v1.1 in different data settings: (a) the full-data scenario, and (b) four low-data scenarios with 100, 300, 500, and 1000 training examples.

\paragraph{Experimental Setup.} We split the official training data into separate training (75\%) and development sets (25\%)\footnote{Again, we use the development set to identify the best performing model across all epochs.} and use the official development set as the test data. 
We evaluate the results by computing the F1 score over the word overlap of the predicted answer and the gold answer. 
The hyperparameters search space is provided in Appendix~\ref{sec:hyperparam}.

\paragraph{Results.}
Table~\ref{tab:results of squad} shows our results of RAFT and \baseline. 
Compared to GLUE, that consists of sentence-level text matching tasks, SQuAD is a more complex task in which the model needs to comprehend a longer text sequence to predict an answer span. 
The increased task difficulty is especially reflected in the low-data scenarios, as the performances of both models are below 25 points when only 100 or 300 annotated examples are available. 
As a result, when there are not enough annotated examples available to learn the task, the use of RAFs instead of GELU is not beneficial for the Transformer model. 
However, we again see that RAFT outperforms the \baseline model as enough training examples become available.

In addition, we observe that tuning RAFs during fine-tuning (RAFT$^\mathrm{full}$) is more beneficial compared to fixing RAFs (RAFT$^\mathrm{fixed}$) when the task is more complex.
Considering our findings on the GLUE benchmark, we conjecture that the task difficulty may play an additional role besides the amount of available training data for the performance of RAFT$^\mathrm{full}$ vs. RAFT$^\mathrm{fixed}$; however, this remains to be investigated in future work. %, there exists different ranges with respect to the available training data where using Transformers with (flexible and fixed) RAFs have an advantage.  

\begin{table}
\resizebox{\columnwidth}{!}{%
\begin{threeparttable}
\begin{tabular}{@{}llllll@{}}
\toprule
             & 100 examples\tnote{1} & 300 examples\tnote{1} &500  examples\tnote{1}       & 1000 examples\tnote{1}       & full data\tnote{2} \\ \midrule
\baseline & \textbf{12.72±1.54} & \textbf{22.11±2.46} & 26.46±1.42  & 34.58±1.68      &    72.33±0.23     \\
RAFT$^\mathrm{full}$      & 11.81±0.95 & 19.49±2.01 & \textbf{26.68±1.91} & \textbf{36.69±1.56} & \textbf{74.45±0.47} \\
RAFT$^\mathrm{fixed}$      & 12.19±1.08 & 19.00±2.68 & 26.27±1.39          &     35.98±1.81   &   74.38±0.25 \\ \bottomrule
\end{tabular}%
\begin{tablenotes}
\footnotesize
\item[1] Results are averaged over ten random seeds: 5309, 202206, 20220602, 2259, 49, 2022, 1046, 622, 320, 53
\item[2] Results are averaged over six random seeds: 5309, 202206, 20220602, 2259, 49, 2022
\end{tablenotes}
\end{threeparttable}
}
\caption{Results of RAFTs and \baseline on SQuAD.}
\label{tab:results of squad}
\end{table}

\section{Analysis}\label{sec:analysis}
% Finally, we provide analysis on RAFT regarding its zero-shot capabilities and the shapes of the learned activation functions after pre-training and fine-tuning. Besides, we evaluate RAFT under parameter-efficient tuning paradigm.
%We further provide analysis on the impact of RAF initialization, the zero-shot capability of RAFT, the learned shapes, and its potential with another parameter efficient fine-tuning method.

\begin{table}[!tb]
\centering
\resizebox{0.8\columnwidth}{!}{%
\begin{tabular}{@{}ccc@{}}
\toprule
 & Validation Loss & \multicolumn{1}{l}{Validation PPL.} \\ \midrule
Identity & Divergent & Divergent \\
RELU & 1.626 & 5.08 \\
GELU & \textbf{1.611} & \textbf{5.00} \\ \bottomrule
\end{tabular}%
}
\caption{Different initializations of RAF.}
\label{tab:init_rafs}
\end{table}

\paragraph{Impact of RAF initialization.}
To investigate how initialization affects the performance of RAFT, we train RAFT models initialized with GELU, RELU, and the identity function.
Other hyperparameters are the same as those in section ~\ref{sec:pretraining}.
Table~\ref{tab:init_rafs} shows the performance of different initialization methods during pre-training. 
As we can see, choosing common activation functions such as ReLU or GELU leads to a similar performance while using the identity function for initialization leads to divergence.  

\paragraph{Zero-shot generalization.} To investigate if the higher performances of RAFT vs \baseline come from overfitting on the in-domain data, we conduct cross-domain zero-shot experiments. 
We use the models that have been fine-tuned on MNLI and SQuAD in the full-data scenario and evaluate them on the same tasks but for different data, namely, SNLI \cite{bowman-etal-2015-large} and TriviaQA \cite{joshi-etal-2017-triviaqa}, respectively.
MNLI and SNLI are both datasets that aim to evaluate natural language inference while SQuAD and TriviaQA contain examples for evaluating reading comprehension in different domains.  
Table~\ref{tab:zero-shot} shows the results of our zero-shot evaluation. 
We observe that the increased flexibility and adaptivity of RAFT does not negatively impact its generalization capabilities.
In fact, both variants of RAFT consistently achieve better performance than the corresponding \baseline model. 
% We therefore conclude that the improved performances of RAFT on GLUE and SQuAD cannot be attributed to overfitting on the in-domain data. 

\begin{table}
\centering
\resizebox{1\columnwidth}{!}{%
\begin{tabular}{@{}cccc@{}}
\toprule
\multirow{2}{*}{} & \multicolumn{1}{l}{\multirow{2}{*}{SNLI}} & \multicolumn{2}{c}{Trivia QA} \\
             & \multicolumn{1}{l}{} & verified-web        & verified-wiki       \\ \midrule
\baseline & 74.22±0.19           & 24.62±1.48          & 21.01±0.75          \\
RAFT$^\mathrm{full}$     & \textbf{74.80±0.29}  & \textbf{25.40±1.84} & 21.50±0.76          \\
RAFT$^\mathrm{fixed}$      & 74.76±0.25           & \textbf{25.40±1.25} & \textbf{21.78±0.87} \\ \bottomrule
\end{tabular}%
}
\caption{Zero-shot performance of \baseline and RAFT. Models evaluated on SNLI are trained on MNLI. Results on TriviaQA are based on models trained on SQuAD.}
\label{tab:zero-shot}
\end{table}

\begin{figure}
\centering
    \begin{subfigure}{0.45\columnwidth}
        \includegraphics[width=\textwidth]{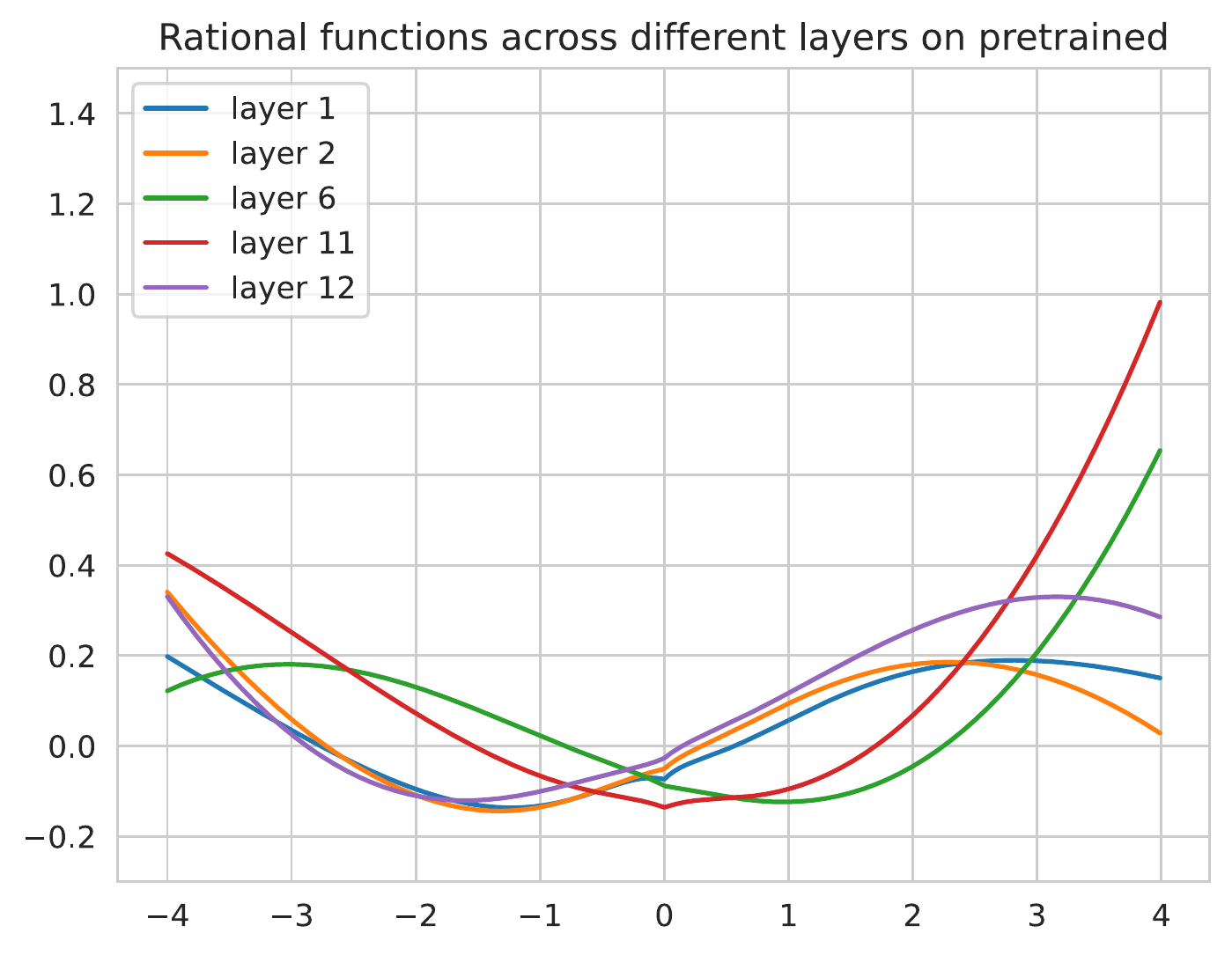}
        \caption{Pre-training}
        \label{fig:rafs-pretrain}
    \end{subfigure}
    \begin{subfigure}{0.45\columnwidth}
        \includegraphics[width=\textwidth]{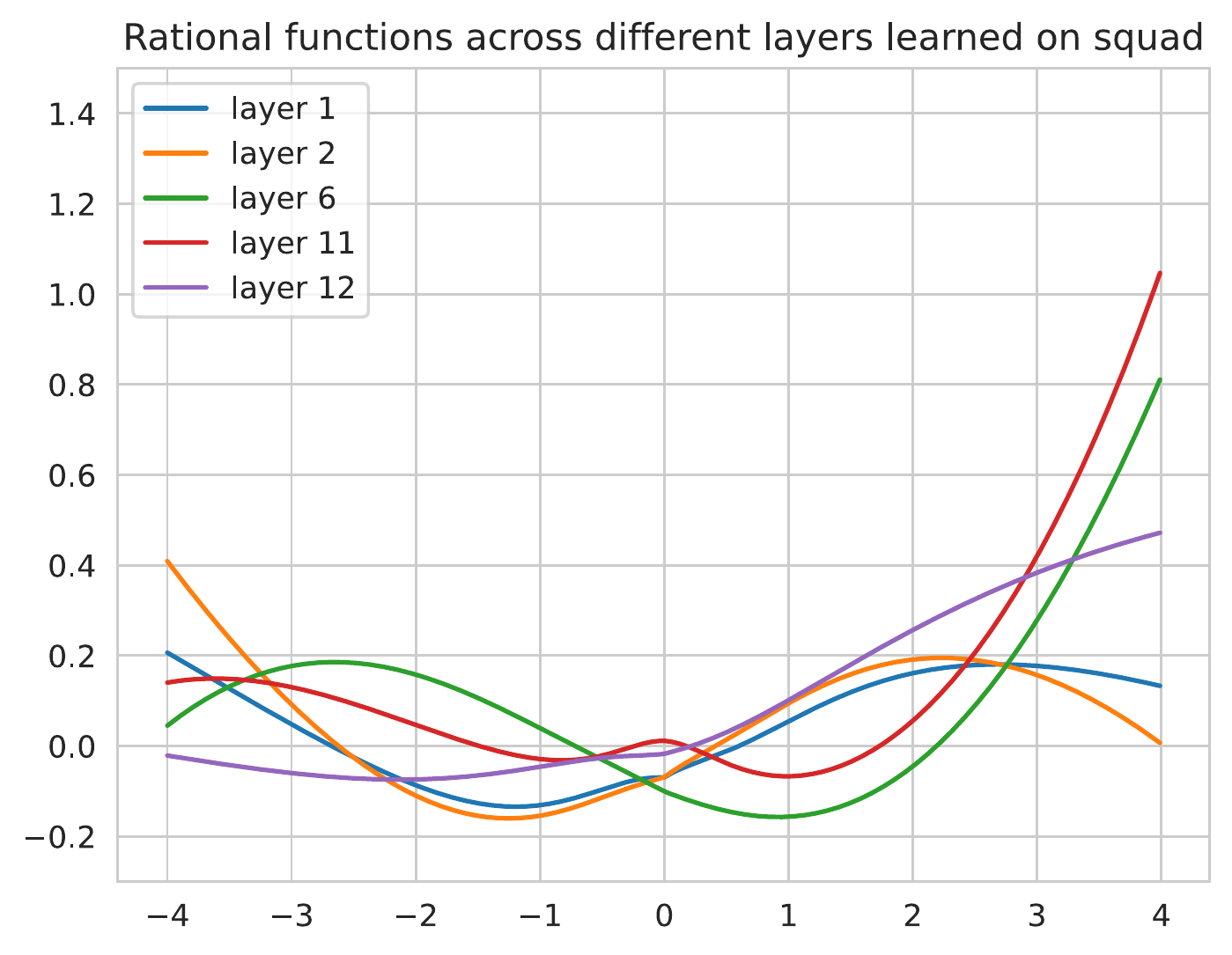} 
        \caption{Fine-tuned on SQuAD}
        \label{fig:rafs-SQuAD}
    \end{subfigure}
    \begin{subfigure}{0.45\columnwidth}
        \includegraphics[width=\textwidth]{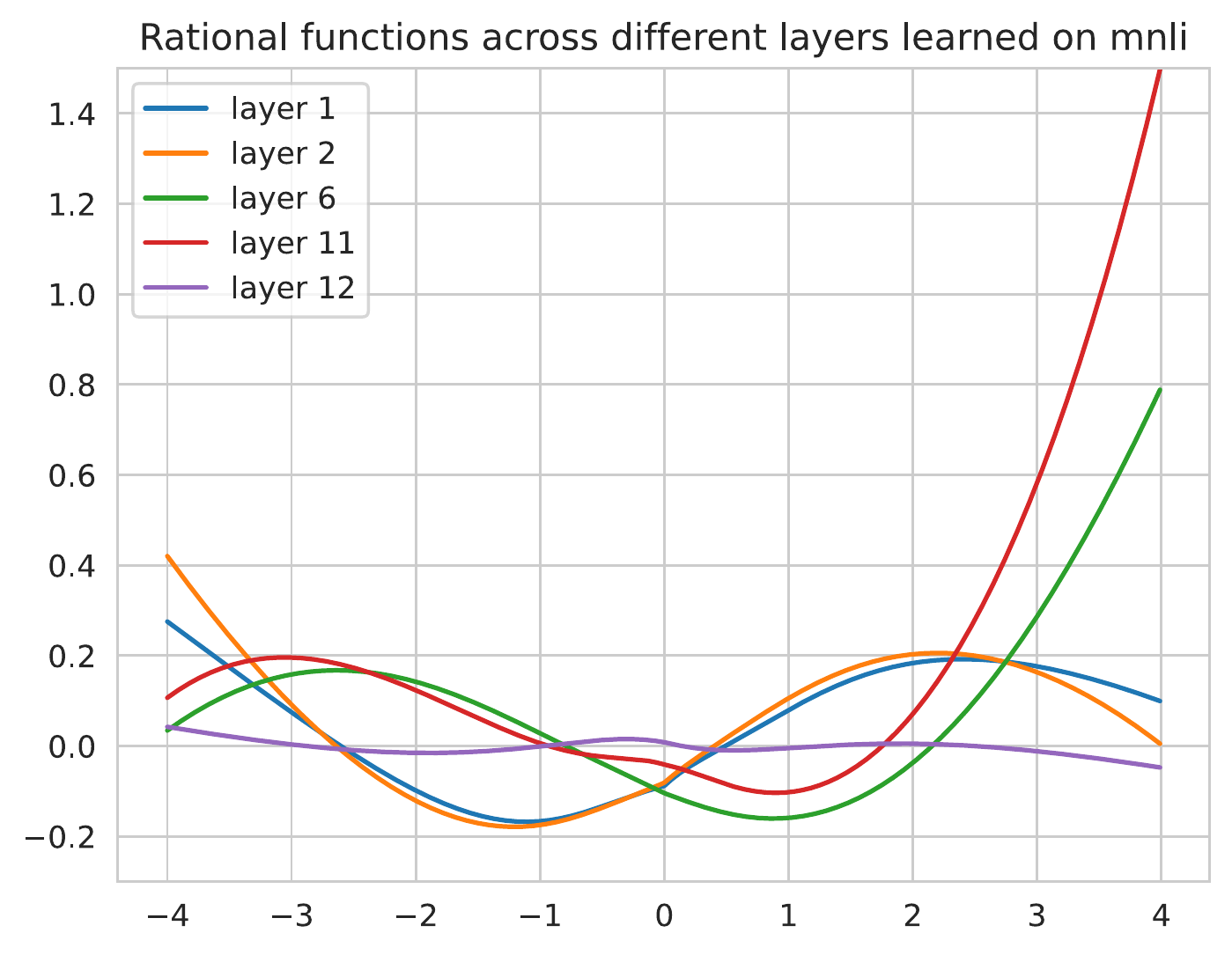}
        \caption{Fine-tuned on MNLI}
        \label{fig:rafs-MNLI}
    \end{subfigure}
    \begin{subfigure}{0.45\columnwidth}
        \includegraphics[width=\textwidth]{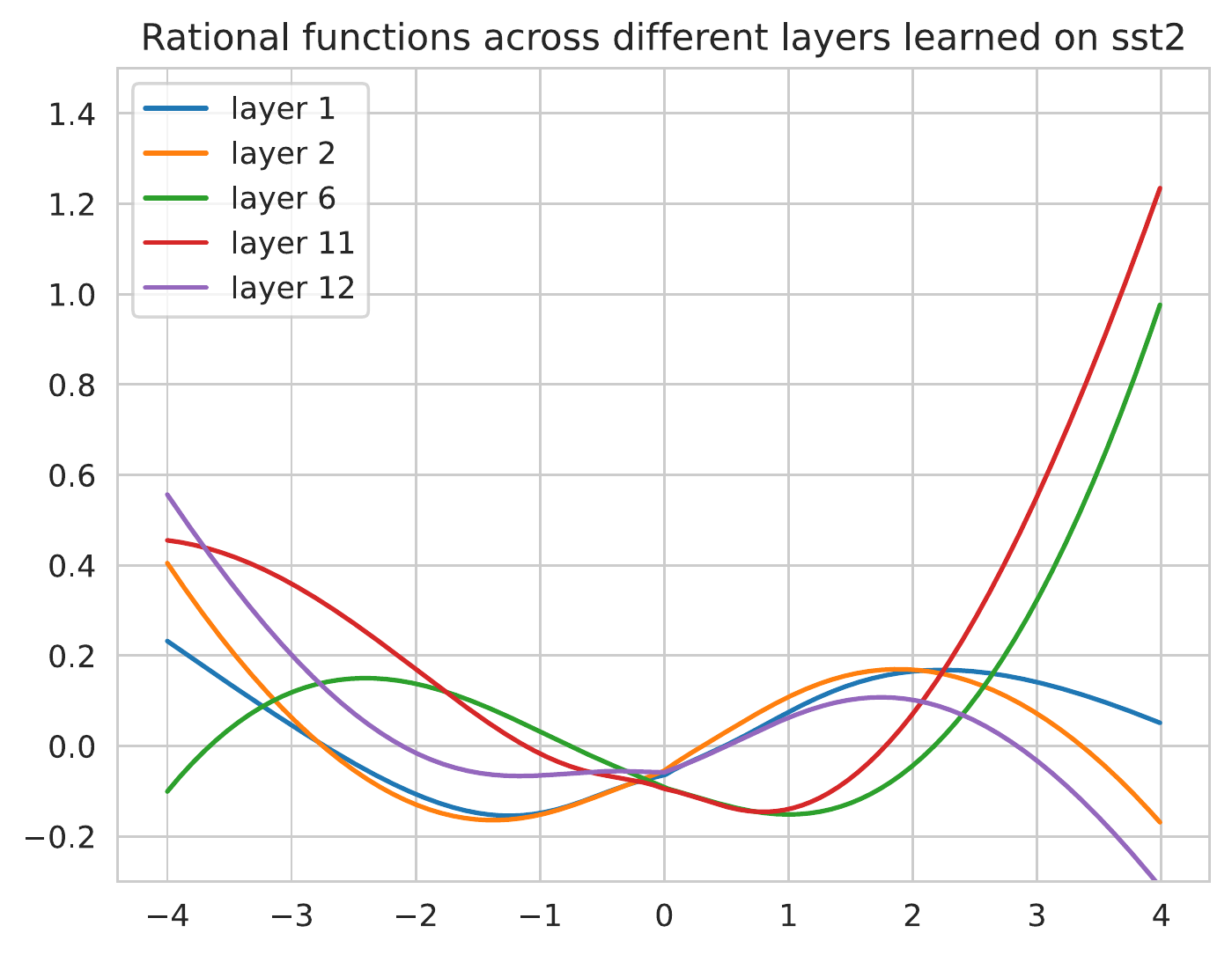} 
        \caption{Fine-tuned on SST2}
        \label{fig:rafs-SST2}
    \end{subfigure}
\caption{Rational activation functions of RAFT$^\mathrm{full}$ among different layers after pre-training and fine-tuning.}
\label{fig:rafs}
\end{figure}

\paragraph{Visualizing learned RAFs.} 
Next, we analyze how the shapes of RAFs change after pre-training and fine-tuning. 
% , and for different tasks.
First, we analyze the learned RAFs in different layers of RAFT after pre-training. 
% to see whether they resemble the common predefined activation functions. 
As shown in Figure~\ref{fig:rafs-pretrain}, rational functions have different shapes across different layers, none of which are similar to GELU, or other commonly used activation functions in Transformers (cf. Table~\ref{tab:activation_functions_models}). 
%Instead, we find some interesting patterns in specific layers, e.g., a similarity to the Swish activation function~\citep{Ramachandran} in layers 6 and 11 (cf. Figure~\ref{fig:rafs}).
This indicates that different layers may need different activation functions to achieve the optimal performance.
Moreover, we see that some features like monotonicity that often are deemed to be good for predefined activation functions are not necessary, which is in line with the findings of the Swish activation function \cite{ramachandran2017searching}.

% \begin{figure}[]
%     \centering
%     \includegraphics[width=0.8\columnwidth]{./images/pretrained.pdf}
%     \caption{Rational activation functions across different layers after pre-training}
%     \label{fig:rf_pre-train}
% \end{figure}

Second, we analyze how the learned RAFs during pre-training change after fine-tuning in RAFT$^\mathrm{full}$.
Figures~\ref{fig:rafs-SQuAD}--\ref{fig:rafs-SST2} show learned RAFs after fine-tuning RAFT$^\mathrm{full}$ on SQuAD, MNLI and SST2 datasets. 
We observe that some of the learned RAFs trained on these three tasks differ from each other and the RAFs after pre-training.
We further see that several RAFs between both tasks have similar shapes but different slopes across many layers.

\begin{figure*}
    \centering
    \includegraphics[width=.3\textwidth]{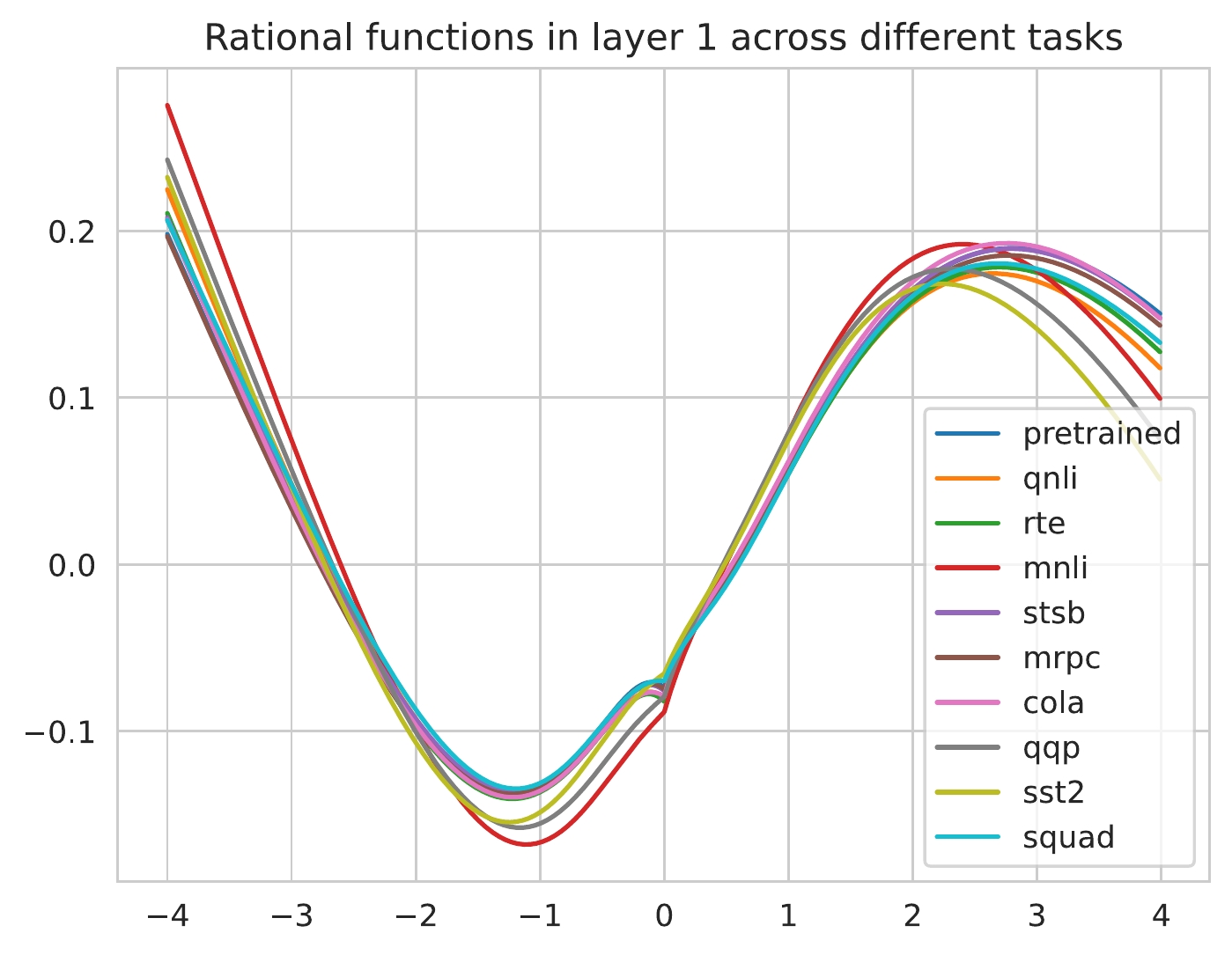} \hfill
    \includegraphics[width=.3\textwidth]{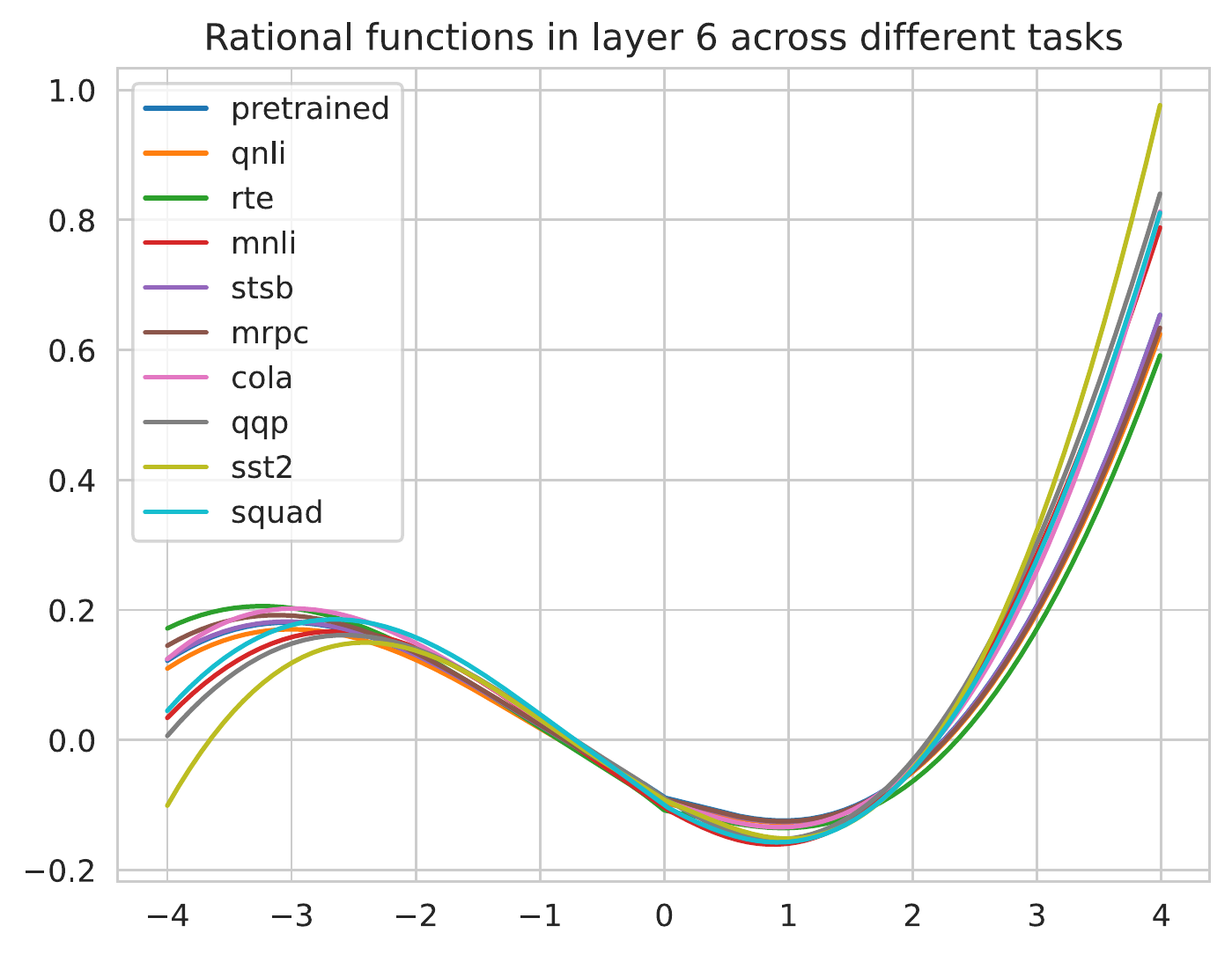} \hfill
    \includegraphics[width=.3\textwidth]{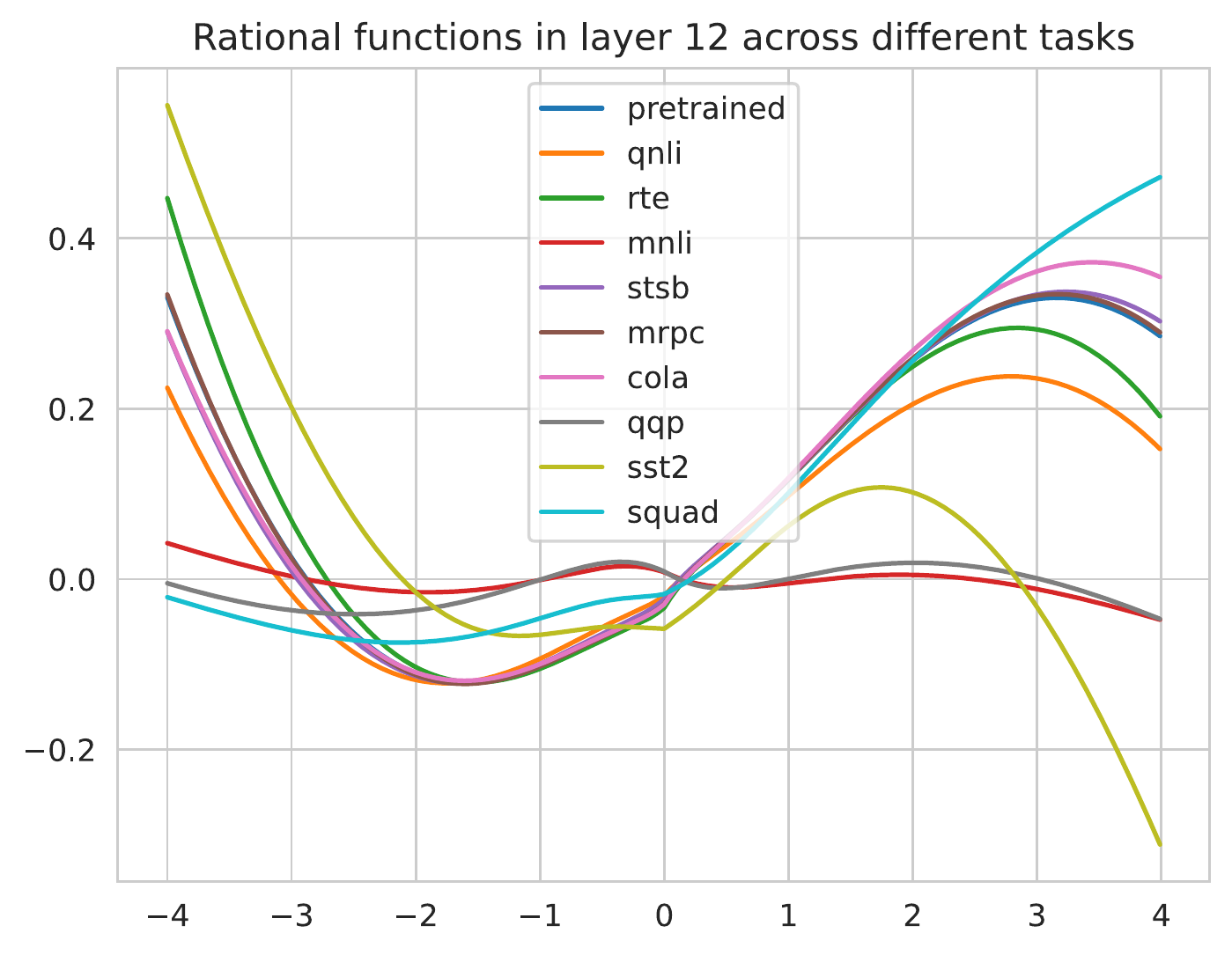}
    \caption{Learned rational activation functions of RAFT$^\mathrm{full}$ in layers 1 (bottom), 6, and 12 (top) among different tasks.}
    \label{fig:rf_tasks}
\end{figure*}

\begin{table*}[!t]
\centering
\resizebox{\textwidth}{!}{%
\begin{threeparttable}
\begin{tabular}{@{}cccccccccc@{}}
\toprule
Model & ColA & SST2 & MRPC & QQP & STSB & MNLI-matched/mismatched & QNLI & RTE & Avg. \\ \midrule
\textit{low data 100 examples\tnote{1}} &  &  &  &  &  &  &  &  &  \\
$BitFit_{\mathrm{\baseline}}$ & 1.44±2.85 & 63.33±9.63 & 68.82±1.74 & 55.49±3.94 & 46.04±24.69 & 32.92±1.33/32.95±1.24 & 51.95±3.50 & \textbf{52.20±2.82} & 45.02 \\
$BitFit_{\mathrm{full}}$ & 4.39±3.41 & \textbf{76.49±1.90} & 74.11±1.04 & \textbf{61.53±3.09} & \textbf{50.41±20.20} & \textbf{33.75±1.38}/33.81±1.30 & \textbf{57.22±6.15} & 50.83±2.74 & \textbf{49.17} \\
$BitFit_{\mathrm{fixed}}$ & \textbf{6.25±3.68} & 75.96±1.24 & \textbf{74.71±0.34} & 61.35±3.42 & 49.91±26.88 & 33.73±1.40/\textbf{34.04±1.71} & 53.19±4.02 & 51.63±2.26 & 48.97 \\
\midrule
\textit{Full data\tnote{1}} & \textit{} & \textit{} & \textit{} & \textit{} & \textit{} & \textit{} & \textit{} & \textit{} & \textit{} \\
$BitFit_{\mathrm{\baseline}}$ & 37.75±1.26 & 87.80±0.67 & 82.94±1.20 & \textbf{81.35±0.13} & 59.29±33.04 & \textbf{71.94±0.38/73.57±0.38} & \textbf{85.38±1.07} & 55.89±1.70 & 70.66 \\
$BitFit_{\mathrm{full}}$ & 38.46±1.37 & 88.19±0.16 & \textbf{86.73±1.00} & 81.03±0.12 & 85.28±0.33 & 70.23±0.41/72.53±0.33 & 80.51±10.75 & \textbf{60.72±1.88} & \textbf{73.74} \\
$BitFit_{\mathrm{fixed}}$ & \textbf{39.96±1.95} & \textbf{88.46±0.28} & 84.91±5.10 & 81.02±0.14 & \textbf{85.55±0.44} & 71.25±0.19/73.26±0.36 & 77.23±14.23 & 60.15±0.90 & 73.53 \\
\bottomrule
\end{tabular}%
\begin{tablenotes}
\item[1] Results are averaged over five random seeds: 5309, 202206, 20220602, 2259, 49
\end{tablenotes}
\end{threeparttable}
}
\caption{Comparison between RAFT and \baseline combined with BitFit.}
\label{tab:bitfit}
\end{table*}

To better understand the behavior of learned RAFs after fine-tuning in different layers on various tasks, we plot RAFs from the same layer together across all tasks.
Figure~\ref{fig:rf_tasks} shows the learned RAFs in layer 1 (the bottom layer), layer 6, and layer 12 (the top layer) after pre-training and fine-tuning on different tasks.
We observe that after fine-tuning, the RAFs in the top layer are more task-specific and change the most, compared to those in bottom layers. 
This is in line with prior work that analyzed the behavior of BERT layers during fine-tuning, which showed that higher layers exhibit more changes compared to lower layers \cite{mosbach2020interplay,merchant2020happens,zhou2021closer}.
Our results confirm this finding from the perspective of learned activation functions. 
It also demonstrates that RAFs can self-adapt to different layers and tasks during fine-tuning.
In addition, an interesting observation is that the output ranges of the RAFs of MNLI and QQP in the top layer are very close to zero.
The output of the FFN layer $Layernorm(\mathrm{FFN}(x)+ x)$ consists of two parts: the feedforward branch $\mathrm{FFN}(x)$ and the skip connection branch $x$.
The very small output of activation functions may indicate that the FFN branch of the top layer does not contribute much to the final model performance on MNLI and QQP and thus could be pruned.
We leave this as future work.

% Overall, we find that RAFT provides a new opportunity to analyze and interpret pre-trained language models from the perspective of learned activation functions. 
% It is also worth investigating the similarities and dissimilarities of layers with similar learned activation functions in terms of the learned embedding space or linguistic properties. 

\paragraph{RAFT$^\mathrm{fixed}$ vs. RAFT$^\mathrm{full}$.} In our experiments on GLUE and SQuAD (Tables~\ref{tab:glue} and \ref{tab:results of squad}), we observe that fixing the RAFs after fine-tuning (RAFT$^\mathrm{fixed}$) often achieves the best or second best performance compared to the full-tuning model (RAFT$^\mathrm{full}$) and \baseline.  
Fine-tuning RAFs results in higher performances when (a) more data is available, i.e., the full-data scenario in GLUE, or (b) the input task is more complex such as in SQuAD. 
% However, the performance gain is minor compared to RAFT$^\mathrm{fixed}$ in our experiments.
We hypothesize that training RAFs during fine-tuning will be more effective when evaluated on more complex tasks and datasets than the ones used this work.

\paragraph{Efficiency comparison between RAFT and \baseline.}
In RAFT, RAFs are polynomial ratios and their coefficients are learned during training, which adds extra computation overhead.
We use RAFs library with CUDA extension to accelerate. 
As shown in Table~\ref{tab:efficiency_com}, RAFT is slower than \baseline during training since RAFs need to be updated (36.8\% slower at pre-training, 14.8\% slower at fine-tuning). However, RAFT is faster when doing inference due to the CUDA implementation (13.8\% faster at pre-training, 3.9\% faster at fine-tuning). 

\begin{table}
\centering
\resizebox{\columnwidth}{!}{%
\begin{threeparttable}
\begin{tabular}{@{}ccccc@{}}
\toprule
steps/second & \multicolumn{2}{c}{Pre-training} & \multicolumn{2}{c}{Fine-tuning} \\ \midrule
\multicolumn{1}{l}{} & Train & Inference & \multicolumn{1}{l}{Train} & \multicolumn{1}{l}{Inference} \\
RAFT & 0.38 & \textbf{3.3} & 12.54 & \textbf{71.05} \\
\baseline & \textbf{0.52} & 2.9 & \textbf{14.4} & 68.38 \\ \bottomrule
\end{tabular}%
\end{threeparttable}
}
\caption{Number of steps per second for training and inference for RAFT and \baseline.}
\label{tab:efficiency_com}
\end{table}

\paragraph{Parameter-efficient fine-tuning with RAFTs.} 
In contrast to fine-tuning all parameters in a pre-trained language model, parameter-efficient tuning techniques that freeze the majority of pre-trained parameters and only fine-tune a small set can be promising alternatives \cite{ding2022delta}.  
One such method is BitFit \cite{zaken2021bitfit} which only updates the bias terms in the Transformer model. 
To investigate the effectiveness of RAFT in a parameter-efficient fine-tuning paradigm, we fine-tune the \baseline and RAFT models with BitFit on the GLUE benchmark. 
We use the same settings as in our previous experiments and test RAFT and \baseline in three configurations in the low-data 100 and full-data scenario: (a) $BitFit_{\mathrm{\baseline}}$ uses BitFit with \baseline, (b) $BitFit_{\mathrm{full}}$ uses BitFit with RAFT$^\mathrm{full}$, and (c) $BitFit_{\mathrm{fixed}}$ uses BitFit with RAFT$^\mathrm{fixed}$. 
As shown in Table~\ref{tab:bitfit}, RAFT-based BitFit achieves higher performance than the \baseline on average in both data settings: $BitFit_{\mathrm{fixed}}$ achieves 3.95 points improvements and $BitFit_{\mathrm{full}}$ gets 4.15 points improvements in the low-data scenario while $BitFit_{\mathrm{fixed}}$ performs better with a 2.87 points boost and $BitFit_{\mathrm{full}}$ performs better with a 3.08 points boost in the full-data scenario. 
It is worth noting that in some tasks, the reported results have a very large standard deviation (e.g., 33.04 for $BitFit_{\mathrm{\baseline}}$ on STSB) due to several random seed runs not converging. 
In our experiments, BitFit is not as stable as fine-tuning the whole model.\\
%\citet{telgarsky2017rational} \shortcite{telgarsky2017rational} demonstrate rational functions are universal approximators which can approximate neural networks.

\begin{figure}
    \centering
    \begin{subfigure}[b]{\columnwidth}
        %  \centering
         \includegraphics[width=\textwidth]{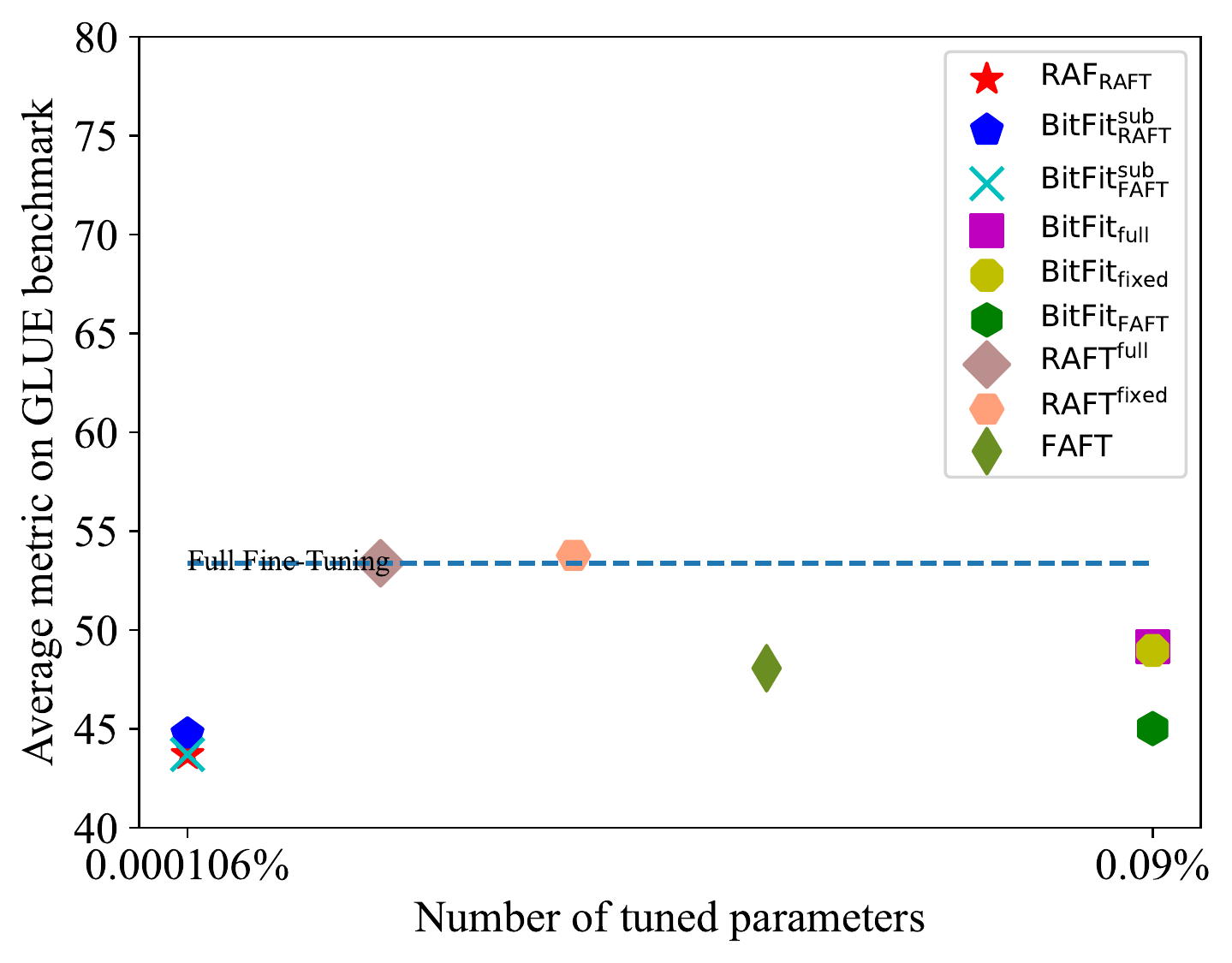}
         \caption{Comparison performance in low-data 100 scenario}
        %  \label{fig:compariosn}
     \end{subfigure}
     %\hifll 
     \begin{subfigure}[b]{\columnwidth}
        %  \centering
         \includegraphics[width=\textwidth]{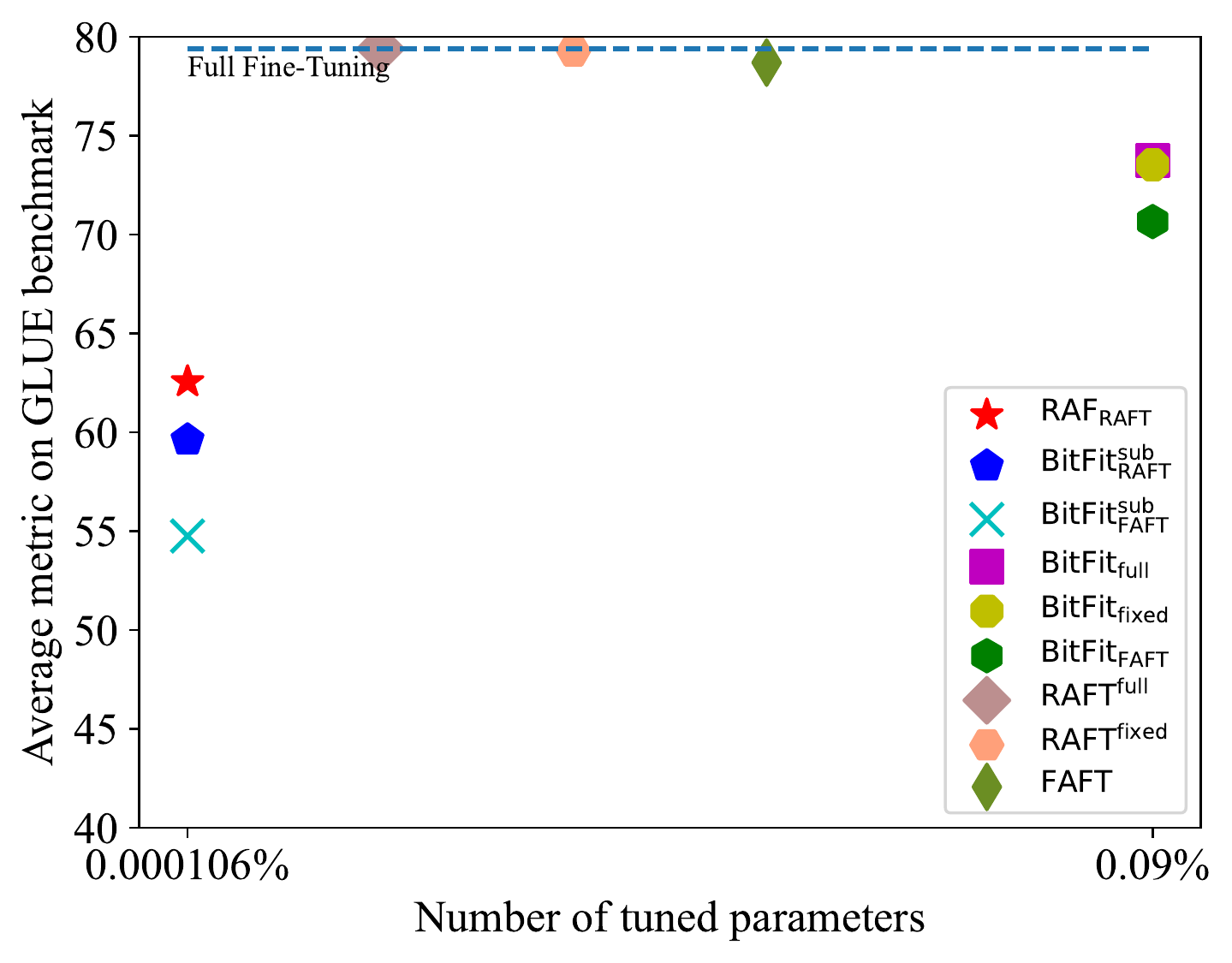}
         \caption{Comparison performance in full-data scenario}
        %  \label{fig:y equals x}
     \end{subfigure}
    \caption{The number of parameters vs. the performance for fine-tuning of RAFT and \baseline.}
    \label{fig:comparison all performance}
\end{figure}
\paragraph{How much can we achieve by only fine-tuning RAFs?}
To see to what extent the model can learn from different tasks by only updating RAFs, we conduct experiments to only tune RAFs on the GLUE benchmark in low- and full-data settings.
We call this setup where only 117\footnote{Including RAF in the pooling layer for classification} parameters of the RAFs are updated during fine-tuning,  $RAF_{\mathrm{RAFT}}$.

For comparison, we tune our models with the BitFit setting using the same amount of parameters, i.e., 117.\footnote{Note that we also update the classification head in all models and experiments.} % I think the heads are always updated
$BitFit_{\mathrm{FAFT}}^{\mathrm{sub}}$ represents tuning the subset of BitFit of \baseline, and $BitFit_{\mathrm{RAFT}}^{\mathrm{sub}}$ represents tuning the subset of BitFit of RAFT.
The result is presented in Appendix~\ref{sec:only_rfs} (Table~\ref{tab:only_raf}).
To compare it from a broader view, we plot Figure~\ref{fig:comparison all performance} based on  Table~\ref{tab:glue}, Table~\ref{tab:bitfit} and Table~\ref{tab:only_raf}. We observe that if only a few annotated examples are available (100 examples), $BitFit_{fixed}$ and $BitFit_{full}$ can achieve better performance than full fine-tuning of \baseline. Only fine-tuning 117 parameters ($BitFit_{\mathrm{FAFT}}^{\mathrm{sub}}$, $BitFit_{\mathrm{RAFT}}^{\mathrm{sub}}$ and $RAF_{\mathrm{RAFT}}$) ---i.e., a negligible number of parameters compared to 110M parameters in \baseline---results in a comparable performance as fine-tuning all the parameters with only a drop of 4.21--6.68 percentage points. 
% However, the performance gap across all three approaches increases in the full data setting (16.72--23.94 drop in pp).
%First, Compared to full fine-tuning, only updating RAFs, i.e., $\sim$0.000106\% of the number of parameters in \baseline, can achieve 81.9\% and 78.8\% of the performance of the low- and full-data setting.
%This experiment shows that only tuning rational activation functions still have strong fitting capacity on downstream tasks.
In the full-data scenario, the performance of BitFit ($BitFit_{full}$, $BitFit_{fixed}$ and $BitFit_{FAFT}$) lags behind full fine-tuning of both models.  
Only tuning RAFs or a subset of BitFit cannot achieve comparable results as well.
However, $RAF_{\mathrm{RAFT}}$ outperforms $BitFit_{\mathrm{FAFT}}^{\mathrm{sub}}$ by 7.8\% and performs better than $BitFit_{\mathrm{RAFT}}^{\mathrm{sub}}$ by 2.94\% in this setting.

\section{Conclusion and Future Work}
In this work, we propose to utilize rational activation functions (RAF) in Transformers to directly learn optimal activation functions from data during pre-training and fine-tuning.
To evaluate the effectiveness of rational activation functions, we pre-trained a Transformer-based language model, namely, RAFT. 
RAFT achieves a lower validation perplexity than \baseline during pre-training. 
Our experimental results show that RAFT performs better than \baseline in general language understanding tasks and reading comprehension tasks across different data size scenarios. 
We further visualize and analyze rational activation functions across different layers and tasks after pre-training and fine-tuning and find that they can substantially vary across different layers and tasks. 
This provides us a new way to analyze and better understand Transformer-based language models.
For instance, we can investigate whether layers with similar rational activation functions encode similar linguistic properties.
We further find that some layers exhibit a close to zero throughput of the rational activation function which indicates that the corresponding feedforward layer does not contribute too much to a model's prediction.
We consider these as our future work.

\section*{Limitations}
\paragraph{Limited training resources.}
This work evaluates the effectiveness of rational activation Transformers using limited GPU resources. 
To provide a fair comparison, we train and release RAF- and GELU-based models for a reduced GPU budget; hence, they are not comparable to publicly available large pre-trained models such as RoBERTa-base etc. 
Still, a fully pre-trained RAFT could be released once more GPU resources are available.
We furthermore note that we use GELU activation functions and the original FFN architecture as our baseline as it is dominantly used in existing models.
% A comparison against models trained using other activation functions such as Swish or as well as GLU variations~ remains to be investigated in future work.
%We only apply RAFs in FFNs of standard Transformers architecture but whether RAFs can still improve the performance of GLU variants~\cite{shazeer2020glu} is not validated.

\paragraph{Societal impact.}
The main focus of this work is the evaluation of trainable activation functions. 
While our visualization of the learned activation functions show that they exhibit substantial differences depending on the downstream task, further analysis is necessary to better understand and interpret the shapes.
Moreover, it is unclear if the additional flexibility of the models may increase their susceptibility towards capturing biases in the data.
At the same time, we conjecture that especially susceptible models could also be used as good indicators to detect such biases.

\section*{Acknowledgements}
We thank Quentin Delfosse for his continued support and valuable advice regarding the existing implementation of rational activation functions. 
We further thank our anonymous reviewers and Stella Biderman, Fengyu Cai, Nils Dycke, Haau-Sing
Li, Andreas R{\"u}ckl{\'e}, Martin Tutek, Kexin Wang, and Neha Warikoo for their fruitful discussions and helpful feedback.
This work has been funded by the German Research Foundation (DFG) as part of the UKP-SQuARE project (grant GU 798/29-1), the German Federal Ministry of Education and Research and the Hessian Ministry of Higher Education, Research, Science and the Arts within their joint support of the National Research Center for Applied Cybersecurity ATHENE and the hessian.AI Service Center.

\bibliography{anthology,custom}

\appendix

% \section{Example Appendix}
\section{Model Architecture} \label{sec:model architecture}
\begin{figure}[!htb]
    \centering
    \includegraphics[width=0.95\columnwidth]{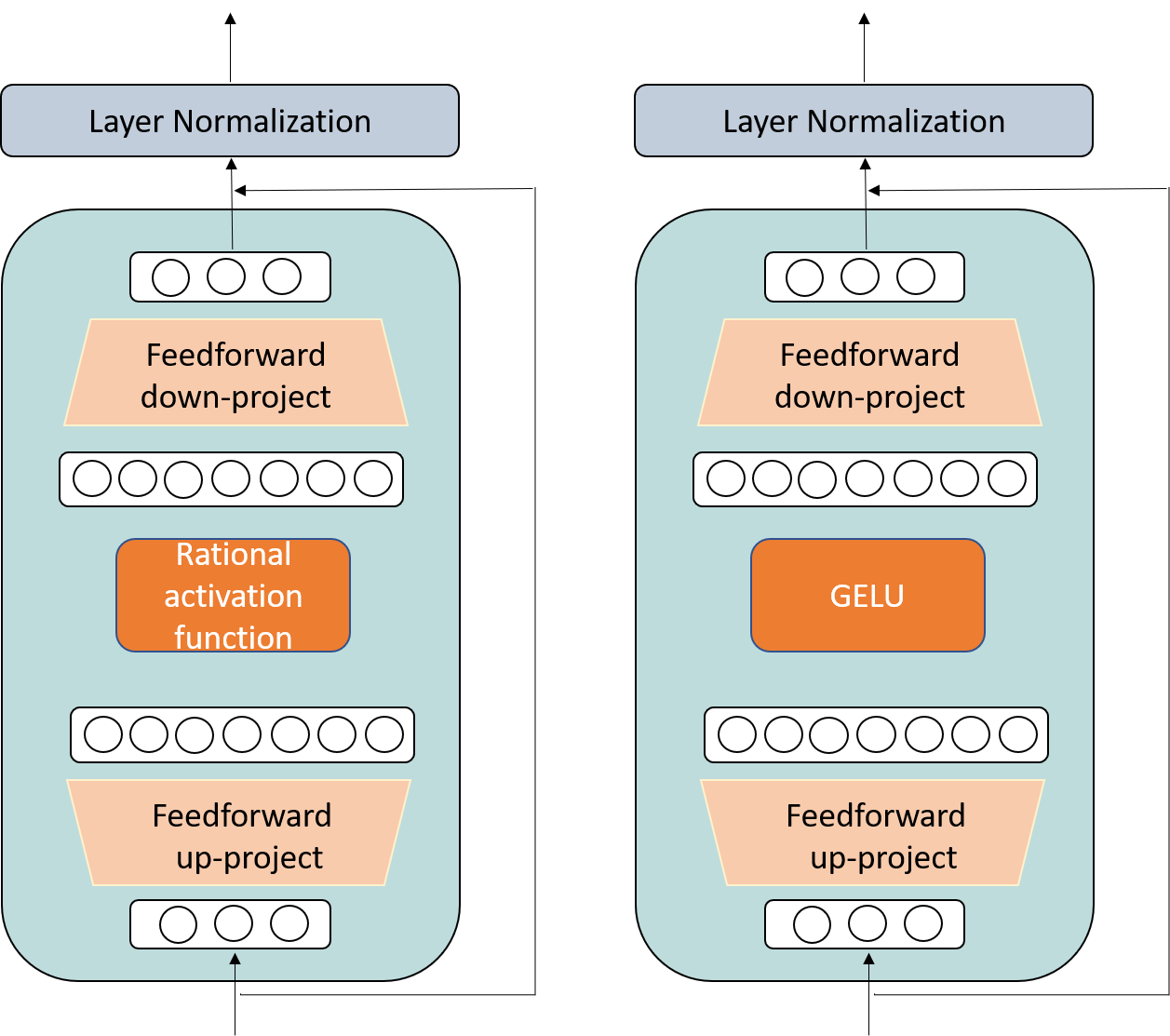}
    \caption{Rational activation function in the feed-forward layer (left) and the vanilla GELU counterpart (right).}
    \label{fig:raft}
\end{figure}

Figure~\ref{fig:raft} shows the difference part of RAFT and \baseline.

\section{Fitting abilities of different degrees of Rational Functions} \label{sec:fitting_abilities}
\begin{figure*}[htb]
    \centering
    \begin{subfigure}{\columnwidth}
       \includegraphics[width=\textwidth]{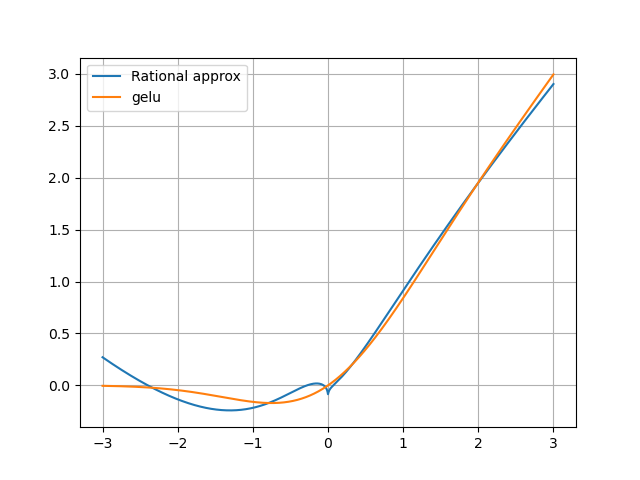}
       \caption{Approximate function with degrees $m=4$ and $n=4$}
    \end{subfigure}
    \begin{subfigure}{\columnwidth}
        \includegraphics[width=\textwidth]{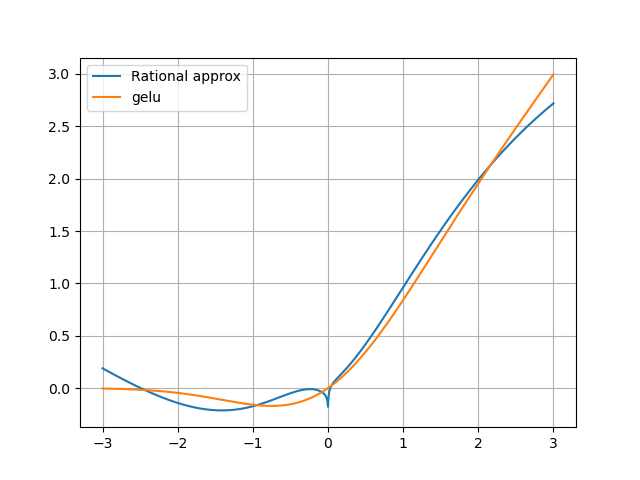}
        \caption{Approximate function with degrees $m=4$ and $n=5$}
    \end{subfigure}
    \begin{subfigure}{\columnwidth}
        \includegraphics[width=\textwidth]{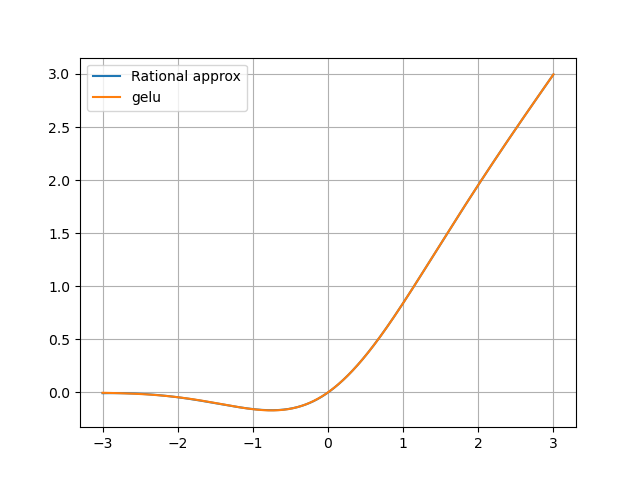}
        \caption{Approximate function with degrees $m=5$ and $n=4$ Rational Function is overlapping with GELU}
    \end{subfigure}
    \begin{subfigure}{\columnwidth}
        \includegraphics[width=\textwidth]{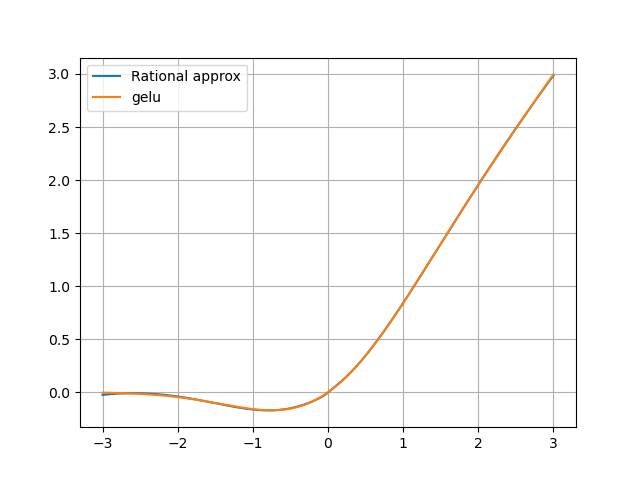}
        \caption{Approximate function with degrees $m=5$ and $n=5$ Rational Function is overlapping with GELU}
    \end{subfigure}
    
    \caption{Approximate Functions of GELU using rational functions}
    \label{fig:approx_gelu}
\end{figure*}

Figure~\ref{fig:approx_gelu} show the approximate functions of GELU using rational functions with different degrees. 
As we can see, when $m=5$ and $n=4$ or $n=5$, rational function fit GELU very well in the same shape.
Finally, it is important to note that rational functions are an universal approximator in a limited range, e.g., [-5,5].
Especially for out-of-bound inputs (i.e., values that are not guaranteed by rational functions), the output of rational functions may result in values very different from the approximated function (e.g., GELU). % as they are not guaranteed by rational functions.  
While pre-training a model from scratch with RAFs does not lead to any problem, directly replacing activation functions in pre-trained models with RAFs only for fine-tuning may lead to divergence due to out-of-bound inputs.

\section{Hyperparameters Tuning}\label{sec:hyperparam}
\subsection{Pre-training}
In our preliminary experiments that some hyperparameter configurations can lead to instability during training due to diverging model updates (e.g., for $lr_\theta\!=\!$7E\nobreakdash-4 and batch size of 2048).  
To stabilize the training without having to rely on a larger warmup phase (e.g., 6\% of the training steps), we instead adopt the DeepNorm \cite{wang2022deepnet} to initialize both models.
DeepNorm stabilizes training by bounding the updates and further scaling the residual branches in Transformers.
Using DeepNorm makes both models, \baseline and RAFT, achieve lower validation loss and leads to a more stable training.

We tune the learning rate $lr_{\theta}$ for model parameters and $lr_{\mathrm{RAF}}$ for RAFs, batch size, warmup steps, and learning rate scheduler as hyperparameters for both models separately.
The hyperparameter search space for pre-training stage is as follows:
\begin{itemize}
    \item Learning rate $lr_{\theta}$ for model parameters: 1E-4, 4E-4, 7E-4, 1E-3
    \item Learning rate $lr_{\mathrm{RAF}}$ for RAFs: 1E-3, 5E-3, 1E-2
    \item Batch size: 2048, 4096
    \item Warmup ratio: 0\%, 1\%, 6\%
\end{itemize}

Some results of hyperparameters tuning are provided in Table~\ref{tab:hyparam-tuning-raft}.

\begin{table}[htb]
\centering
\resizebox{\columnwidth}{!}{%
\begin{tabular}{@{}ccccc@{}}
\toprule
 &  $lr_{\theta}$ & $lr_{\mathrm{RAF}}$ & Batch Size & Validation Loss \\ \midrule
RAFT & 1E-4 & 0.005 & 2048 & 2.217 \\
RAFT & 4E-4 & 0.005 & 2048 & 1.808 \\
RAFT & 7E-4 & 0.005 & 4096 & 1.732 \\
\textbf{RAFT} & \textbf{7E-4} & \textbf{0.005} & \textbf{4096} & \textbf{1.611} \\
RAFT & 1E-3 & 0.005 & 4096 & 1.638 \\ \bottomrule
\end{tabular}%
}
\caption{Part of Hyperparameters Tuning Results of RAFT}
\label{tab:hyparam-tuning-raft}
\end{table}

Table~\ref{tab:pre-train hyperparameters} shows final hyperparameters we used for pre-training RAFT and \baseline.

\begin{table}
\centering
\resizebox{0.85\columnwidth}{!}{%
\begin{tabular}{@{}ccc@{}}
\toprule
Hyperparameters & \baseline & RAFT \\ \midrule
Peak $lr_{\theta}$ & 7E-4 & 7E-4 \\
Peak $lr_{\mathrm{RAF}}$ & n/a & 5E-3 \\
\multicolumn{1}{l}{Learning rate decay} & linear & constant \\
\multicolumn{1}{l}{Gradient clipping} & 0 & 0 \\
Batch size & 4096 & 4096 \\
Sequence length & 128 & 128 \\
Adam\_beta1 & 0.9 & 0.9 \\
Adam\_beta2 & 0.98 & 0.98 \\
Attention dropout & 0.1 & 0.1 \\
Warmup ratio & 1\% & 1\% \\
Training steps & 23k & 23k \\ \bottomrule
\end{tabular}%
}
\caption{Hyperparameters for pre-training RAFT and \baseline}
\label{tab:pre-train hyperparameters}
\end{table}

% Figure~\ref{fig:valid_loss} show the validation loss of RAFT and FAFT.
% \begin{figure}
% % \resizebox{0.5\columnwidth}{}{
%     \centering
%     \includegraphics[width=\columnwidth]{images/validation_ss.png}
%     \caption{Validation loss of RAFT and FAFT}
%     \label{fig:valid_loss}
%     % }
% \end{figure}

\subsection{Fine-tuning}
The hyperparameters search space for GLUE during fine-tuning stage is as follows:
\begin{itemize}
    \item $lr_{\theta}$: 2E-5, 5E-5
    \item $lr_{\mathrm{RAF}}$: 1E-4, 5E-4, 1E-3, 5E-3
    \item Batch size: 32
    \item Weight decay: 0.1
    \item Number of epochs: 3, 10, 20
\end{itemize}
We further tune the learning rates and number of training epochs for RAFT and \baseline separately on a single random seed.
For our low-data experiments we fix the number of training epochs to 20 and use early stopping with a patience of 10 epochs. 
For our full-data experiments, we train the large datasets (QQP, MNLI, and QNLI) for 3 epochs and the others for 10 epochs.

The hyperparameters search space for SQuAD during fine-tuning is as below:
\begin{itemize}
    \item $lr_{\theta}$: 2E-5, 5E-5, 1E-4
    \item $lr_{\mathrm{RAF}}$: 1E-4, 5E-4, 1E-3, 5E-3
    \item Batch size: 32
    \item Weight decay: 0.1
    \item Number of epochs: 10, 20
\end{itemize}
For our experiments, we fine-tune both models with their best performing $lr_\theta\!=\!$1E\nobreakdash-4 for 10 epochs in the full-data scenario and 20 epochs in the low-data scenario. 

The hyperparameters search space for BitFit is as below:
\begin{itemize}
    \item Learning rate $lr_{\theta}$ for model parameters: 5E-5, 1E-3, 5E-3, 1E-2 
    \item Learning rate $lr_{\mathrm{RAF}}$ for RAFs: 1E-3, 5E-3, 1E-2
    \item Batch size: 32
    \item Training epochs: 3, 10, 20 epochs
\end{itemize}
 
We use 3 training epochs for large dataset(QQP, MNLI, QNLI), 10 epochs for other datasets and 20 epochs for low-resource scenarios. Both models can converge in the above settings.

\section{Data Statistics} \label{sec:data}

\paragraph{GLUE} is a collection of nine different language understanding tasks: CoLA \cite{warstadt2018neural}, SST2 \cite{socher2013recursive}, MRPC \cite{dolan2005automatically}, QQP \footnote{\url{https://quoradata.quora.com/First-Quora-Dataset-Release-Question-Pairs}}, STSB \cite{cer2017semeval}, MNLI \cite{N18-1101}, RTE \cite{dagan2005pascal}, and WNLI \cite{levesque2012winograd}. 
We exclude WNLI due to the adversarial nature of its development set and the still unbeaten majority vote upper bound.\footnote{Cf. (12) in \url{https://gluebenchmark.com/faq}}

Table~\ref{tab:statistics on glue} show data statistics of GLUE benchmark.
\begin{table*}
\resizebox{\textwidth}{!}{%
\begin{tabular}{@{}ccccccccc@{}}
\toprule
Task          & CoLA           & SST2   & MRPC    & QQP     & STSB                  & MNLI-matched/mismatched & QNLI    & RTE   \\ \midrule
$|$Train$|$ & 8,551          & 67,349 & 3,668   & 363,846 & 5,749                 & 392,702                 & 104,743 & 2,490 \\
$|$Dev$|$   & 1,043          & 872    & 408     & 40,430  & 1,500                 & 9,815/9,832             & 5,463   & 277   \\
Metric        & Matthews corr. & acc.   & acc./F1 & acc./F1 & Person/Spearman corr. & acc.                    & acc.    & acc.  \\ \bottomrule
\end{tabular}%
}
\caption{Dataset statistics of the GLUE benchmark}
\label{tab:statistics on glue}
\end{table*}

\paragraph{SQuAD} is a reading comprehension task where each example consists of a question, a context, and the respective span from the context that answers the question.
Table~\ref{tab:statistics on SQuAD} show data statistics of SQuAD.

\begin{table*}[htb]
\centering
\begin{tabular}{@{}llll@{}}
\toprule
                              & $|$Train$|$          & $|$Dev$|$    & $|$Test$|$           \\ \midrule
\multicolumn{1}{c}{SQuAD v1.1} & \multicolumn{1}{c}{66,236} & \multicolumn{1}{c}{21,530} & \multicolumn{1}{c}{10,789} \\ \bottomrule
\end{tabular}%
\caption{Statistics of SQuAD: the official training dataset is split into training and development sets, and the official development dataset is used as the test data.}
\label{tab:statistics on SQuAD}
\end{table*}

\section{Learned RAFs during pre-training and after fine-tuning}\label{sec:appendix:e}
\begin{figure*}[htb]
    \centering
    \includegraphics[width=\columnwidth]{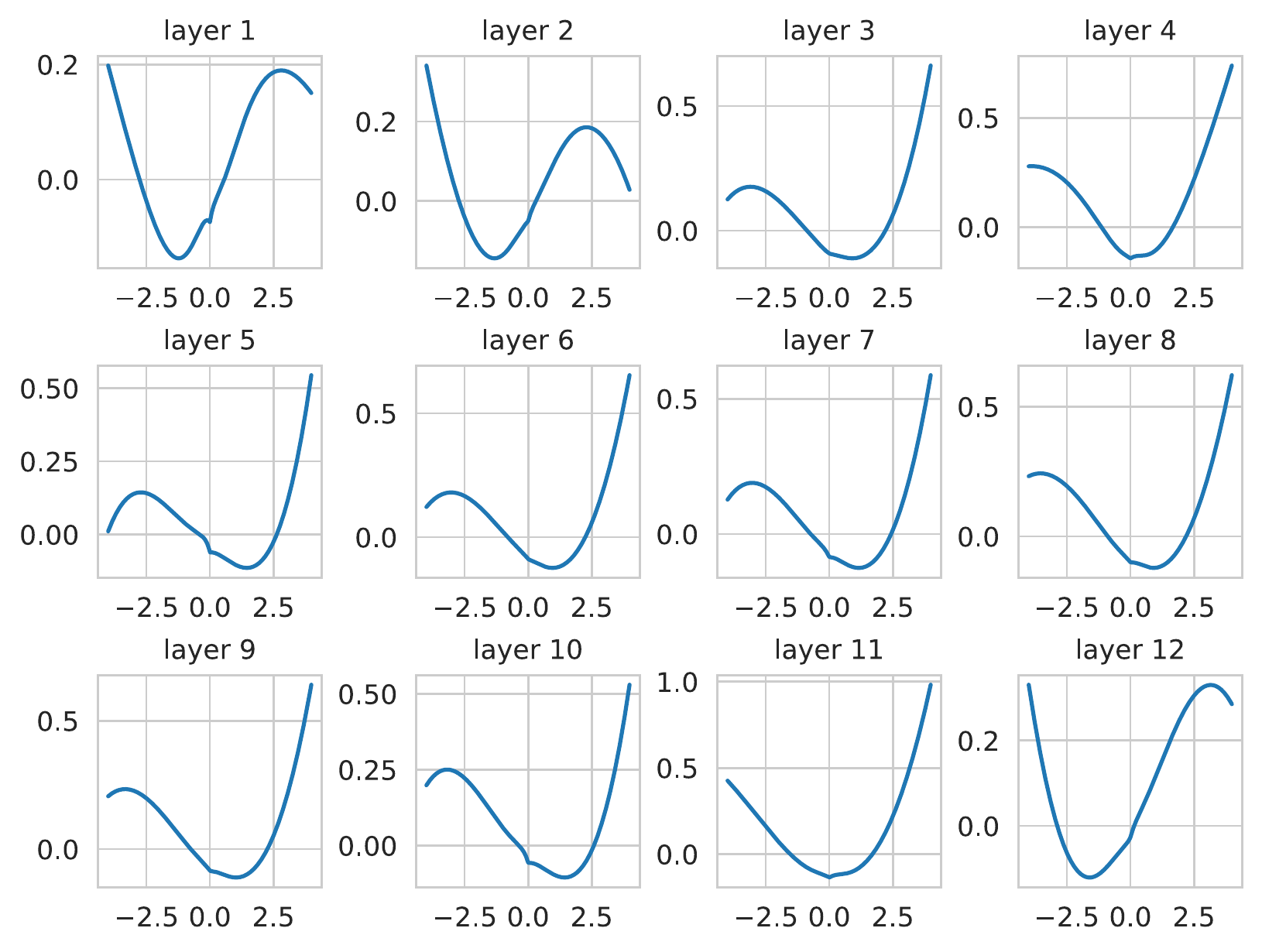}
    \caption{Learned RAFs of different layers after pre-training}
    \label{fig:rfs-after-pretraining}
\end{figure*}

\begin{figure*}[htb]
       \includegraphics[width=0.3\textwidth]{./images/layer_0.pdf} \hfill
       \includegraphics[width=0.3\textwidth]{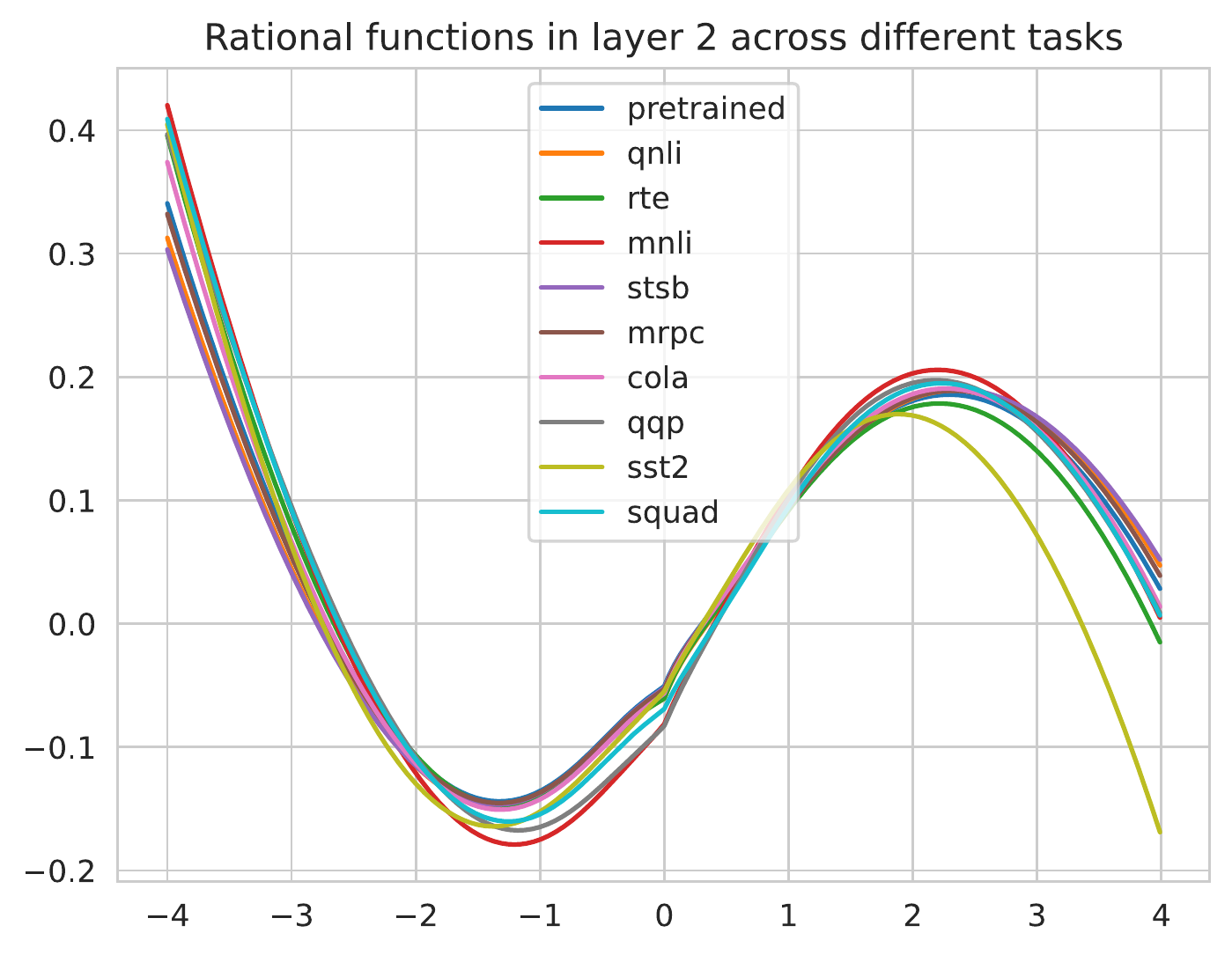} \hfill
       \includegraphics[width=0.3\textwidth]{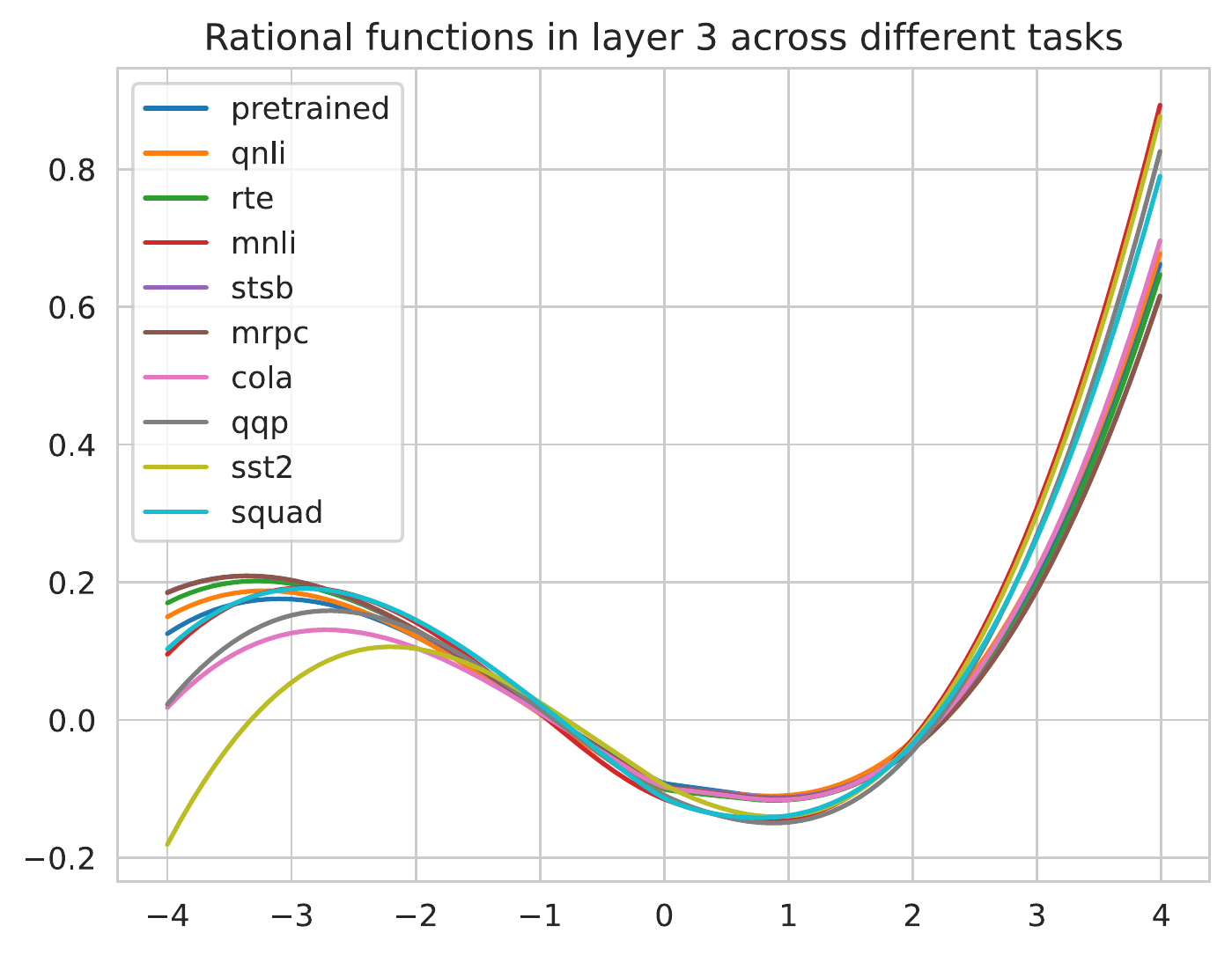}
       
       \includegraphics[width=0.3\textwidth]{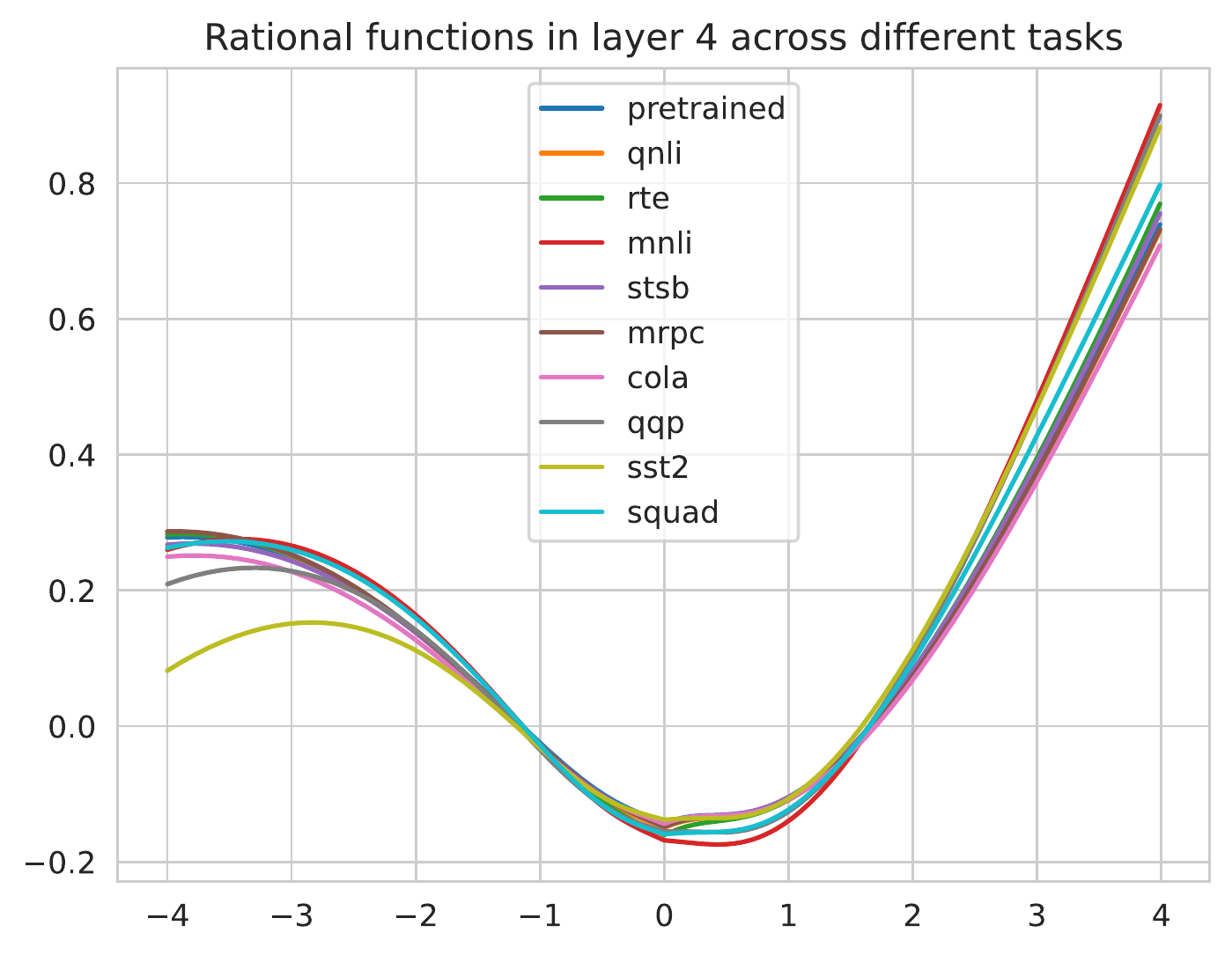} \hfill
       \includegraphics[width=0.3\textwidth]{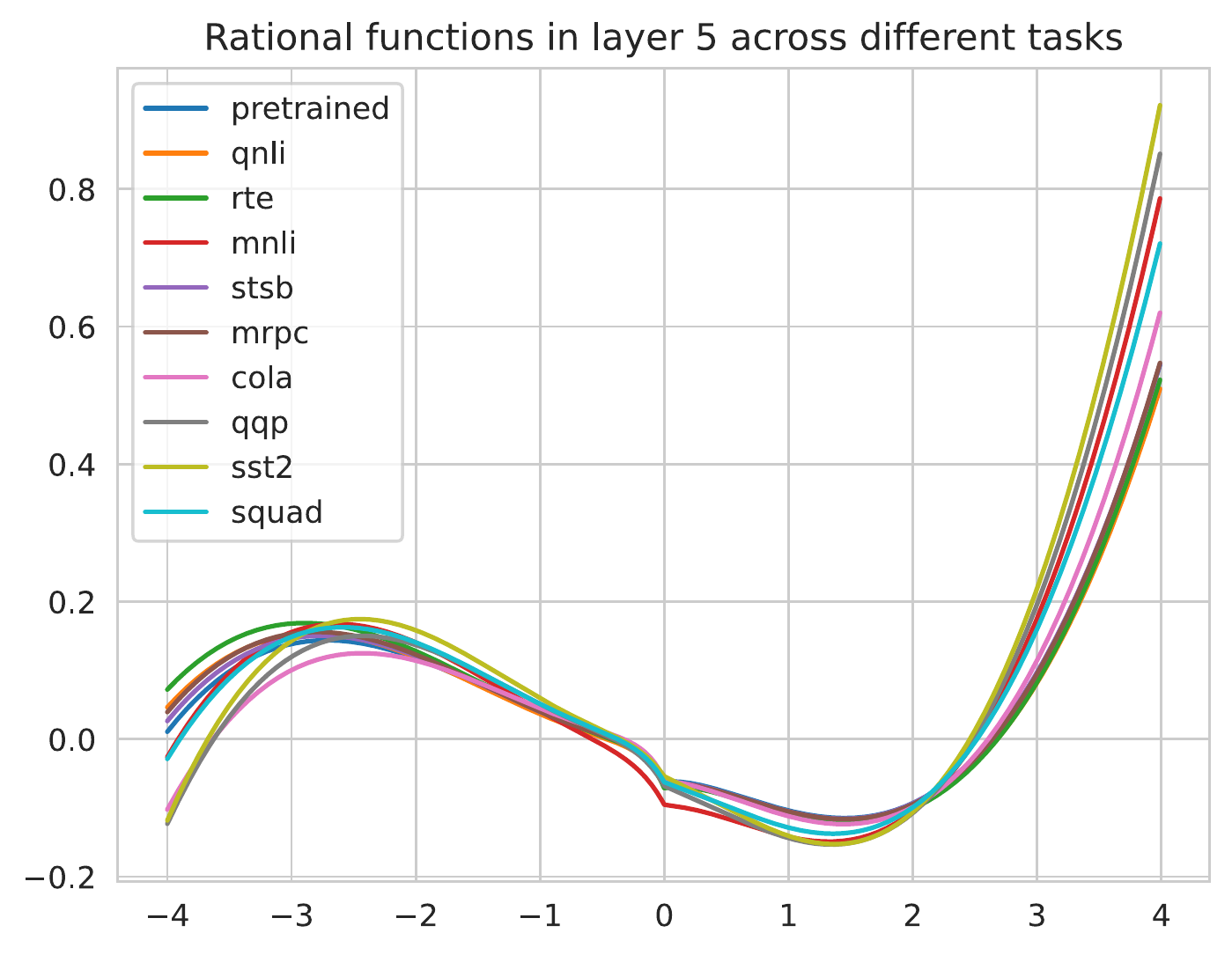} \hfill
       \includegraphics[width=0.3\textwidth]{./images/layer_5.pdf}
       
       \includegraphics[width=0.3\textwidth]{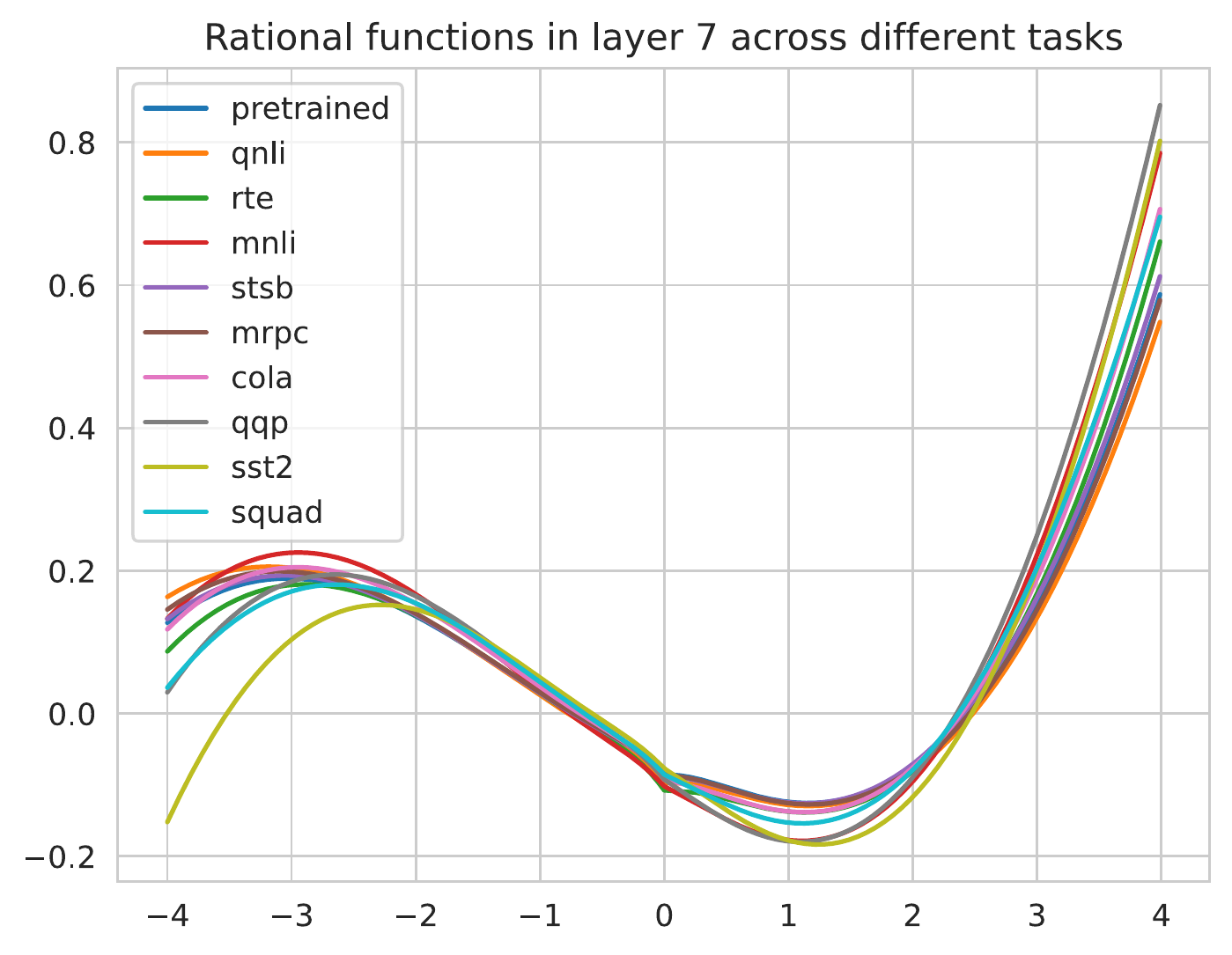} \hfill
       \includegraphics[width=0.3\textwidth]{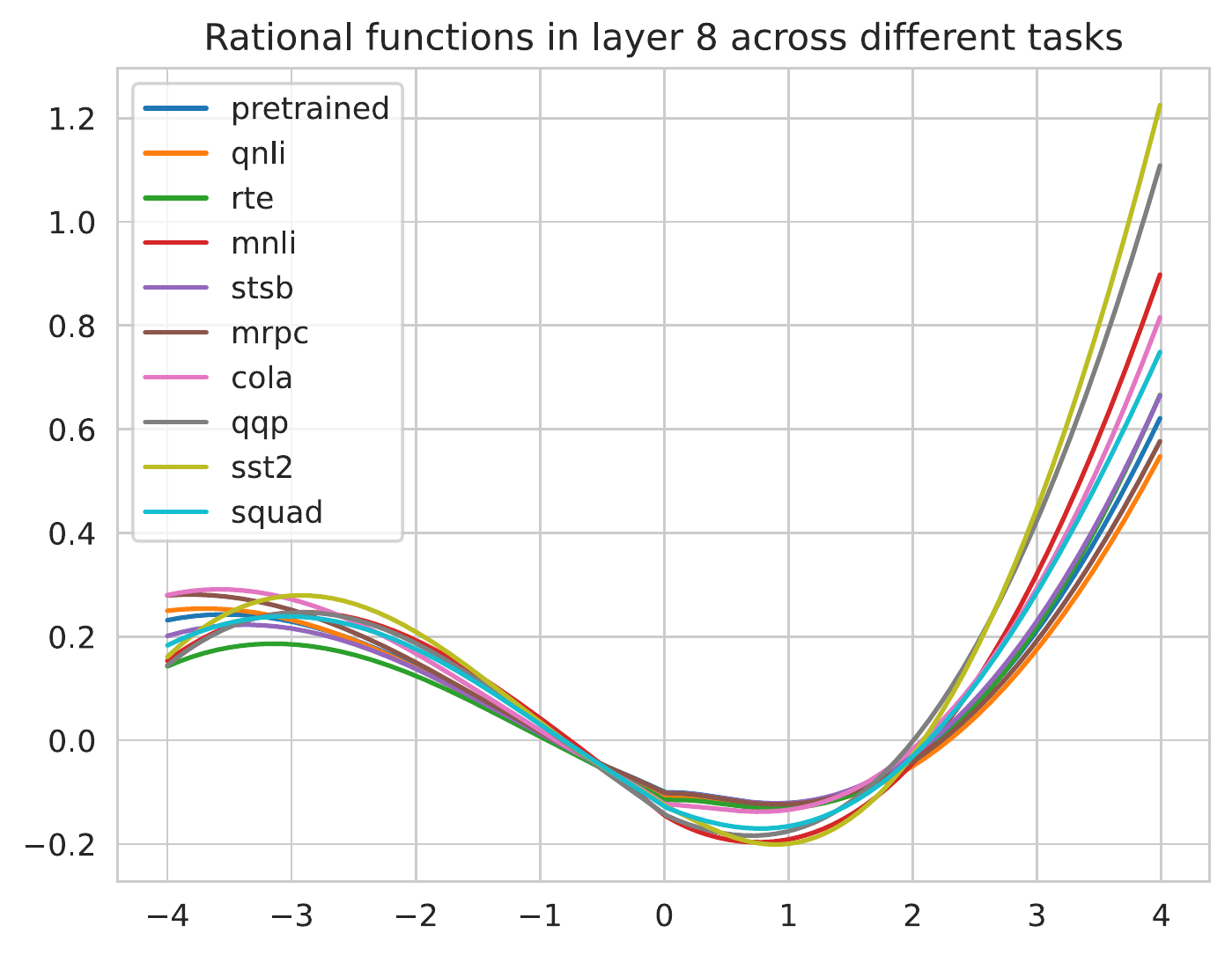} \hfill
       \includegraphics[width=0.3\textwidth]{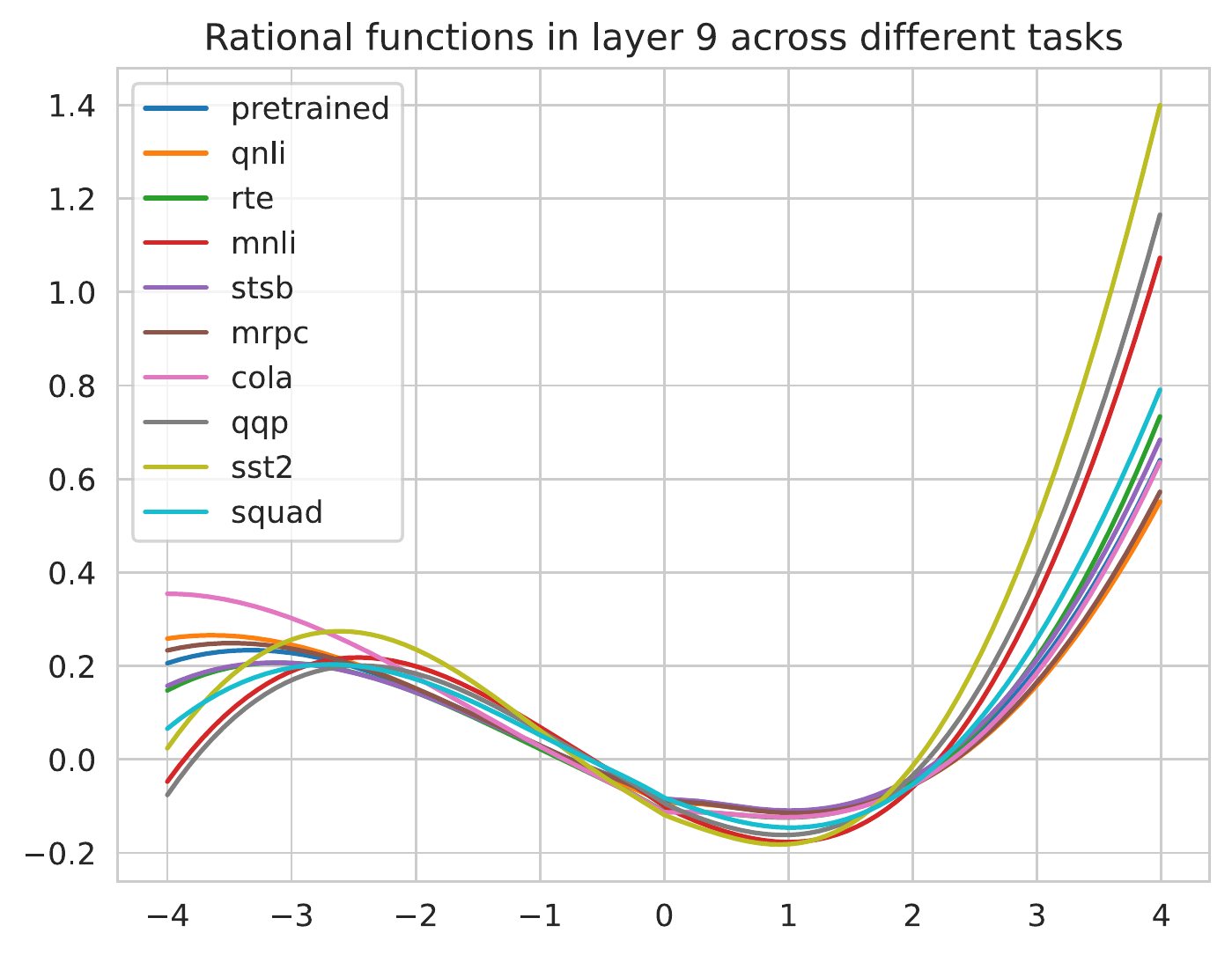}
       
       \includegraphics[width=0.3\textwidth]{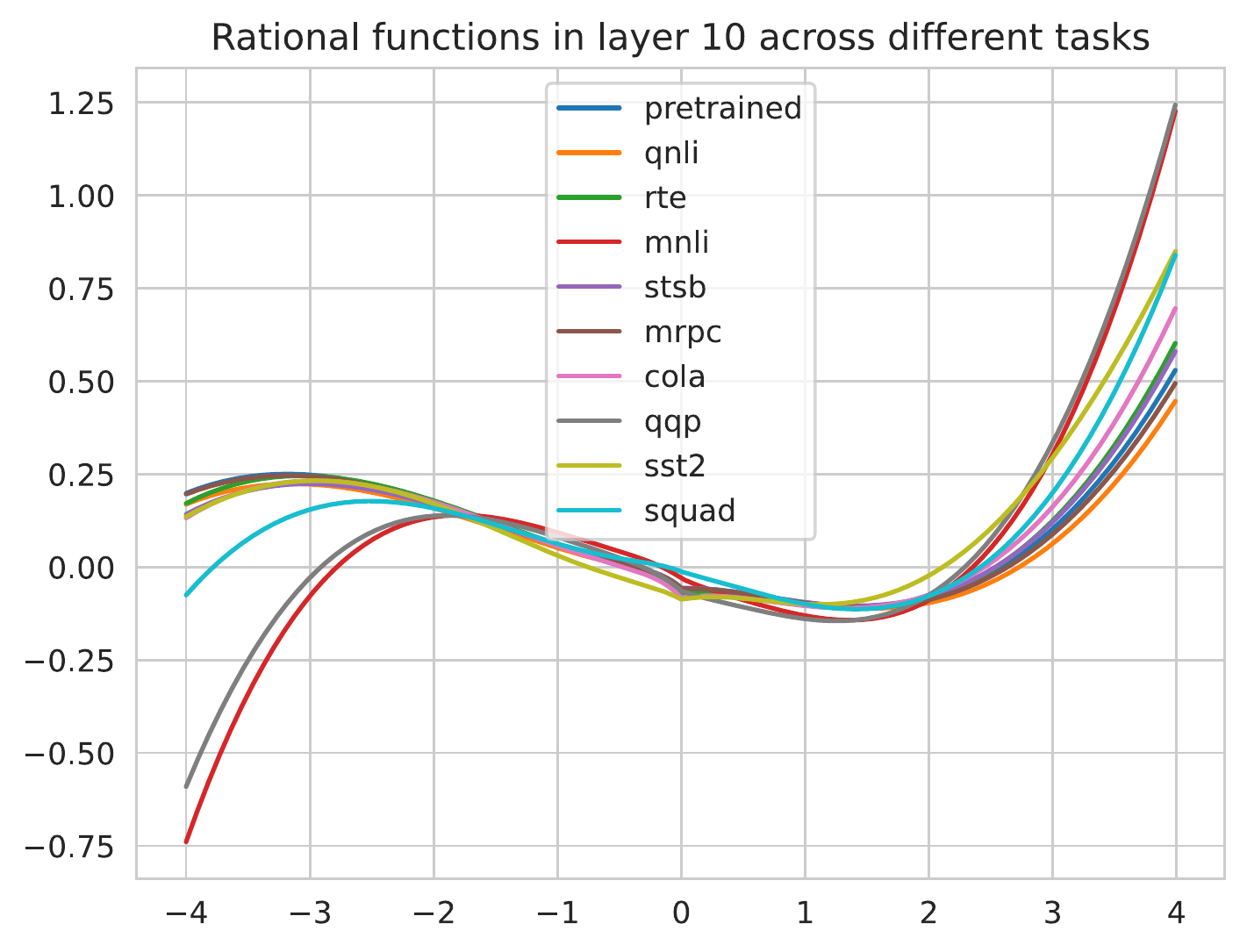} \hfill
       \includegraphics[width=0.3\textwidth]{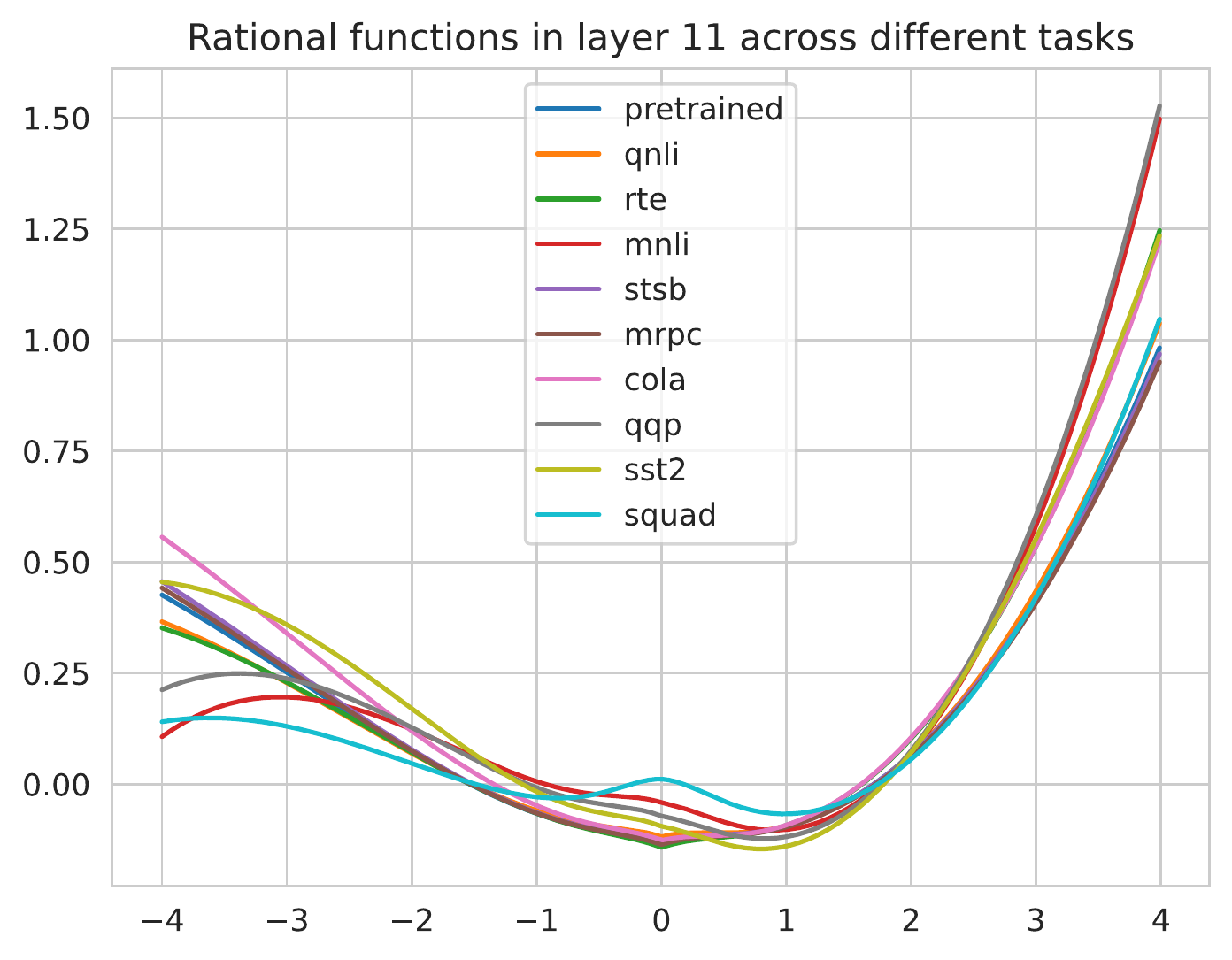} \hfill
       \includegraphics[width=0.3\textwidth]{./images/layer_11.pdf}
       
    \caption{Learned RAFs in 12 layers across different tasks after fine-tuning}
    \label{fig:learned-rf-fine-tuning}
\end{figure*}

Figure~\ref{fig:rfs-after-pretraining} and Figure~\ref{fig:learned-rf-fine-tuning} show learned RAFs in 12 layers after pre-training and fine-tuning on different tasks, respectively.

\section{Results of only tuning RAFs} \label{sec:only_rfs}
Table~\ref{tab:only_raf} shows comparison results between only tuning RAFs and BitFit with the same parameters with RAFT and \baseline.

\begin{table*}[htb]
\resizebox{\textwidth}{!}{%
\begin{threeparttable}
\begin{tabular}{@{}cccccccccc@{}}
\toprule
Model & ColA & SST2 & MRPC & QQP & STSB & MNLI-matched/mismatched & QNLI & RTE & Avg. \\ \midrule
\textit{low data 100 examples\tnote{1}} &  &  &  &  &  &  &  &  &  \\
$BitFit_{\mathrm{\baseline}}^{\mathrm{sub}}$ & 1.49±1.87 & 62.82±7.56 & \textbf{74.80±0.00} & 52.57±3.83 & 14.71±7.21 & 32.73±1.41/32.76±1.30 & 49.77±0.40 & \textbf{50.83±1.86} & 41.39 \\
$BitFit_{\mathrm{RAFT}}^{\mathrm{sub}}$ & 2.45±3.58 & 72.34±3.41 & 74.67±0.68 & \textbf{55.61±2.35} & \textbf{23.99±10.41} & \textbf{35.32±0.67}/35.66±1.05 & 51.08±0.71 & 51.70±1.85 & \textbf{44.75} \\
$RAF_{\mathrm{RAFT}}$ & \textbf{4.33±3.02} & \textbf{72.91±2.82} & 74.47±0.88 & 51.92±5.03 & 17.27±10.60 & 35.24±0.61/\textbf{35.69±0.92} & 51.12±0.48 & 50.47±1.63 & 43.71 \\
\midrule
\textit{Full data\tnote{1}} & \textit{} & \textit{} & \textit{} & \textit{} & \textit{} & \textit{} & \textit{} & \textit{} & \textit{} \\
$BitFit_{\mathrm{\baseline}}^{\mathrm{sub}}$ & 6.61±7.08 & 79.52±0.52 & 71.32±0.22 & 70.48±0.66 & 37.33±5.70 & 53.33±1.13/55.30±0.75 & 64.04±2.03 & 54.88±1.42 & 54.76 \\
$BitFit_{\mathrm{RAFT}}^{\mathrm{sub}}$ & 8.78±5.54 & \textbf{82.02±0.57} & 71.76±0.77 & 70.88±1.17 & 71.40±0.52 & 51.57±0.54/53.27±1.20 & 69.87±1.20 & \textbf{57.04±1.19} & 59.62 \\
$RAF_{\mathrm{RAFT}}$ & \textbf{9.71±12.04} & 81.70±0.12 & \textbf{74.81±3.09} & \textbf{73.57±0.48} & \textbf{80.79±0.60} & \textbf{57.34±0.19/60.69±0.51} & 67.89±8.64 & 56.53±1.83 & \textbf{62.56} \\
\bottomrule
\end{tabular}%
\begin{tablenotes}
\item[1] Results are averaged over five random seeds: 5309, 202206, 20220602, 2259, 49
\end{tablenotes}
\end{threeparttable}

}
\caption{Comparison between fine-tuning RAFs and a subset of 117 BitFit parameters with RAFT and \baseline.}
\label{tab:only_raf}
\end{table*}

\end{document}